\pdfoutput=1

\documentclass[11pt]{article}

\usepackage{acl}

\usepackage{times}
\usepackage{latexsym}
\usepackage{graphicx}
\usepackage{multirow}
\usepackage{colortbl}

\usepackage[T1]{fontenc}

\usepackage[utf8]{inputenc}
\usepackage{url}
\usepackage{microtype}
\usepackage{amssymb}
\usepackage{makecell}
\usepackage{amsmath} 
\usepackage{pifont}

\usepackage{inconsolata}

\usepackage{multirow}
\usepackage{booktabs}
\usepackage{algorithm}
\usepackage{algorithmic}

\title{DAFNet: Dynamic Auxiliary Fusion for Sequential Model Editing in Large Language Models}

\author{Taolin Zhang$^{1}$
\thanks{\ \ T. Zhang, Q. Chen and D. Li contribute equally to this work.}, 
Qizhou Chen$^{2,1}$\footnotemark[1], Dongyang Li$^{2,1}$\footnotemark[1], Chengyu Wang$^{1}$\thanks{\ \ C. Wang and X. He are co-corresponding authors.}, \textbf{Xiaofeng He}$^{2}$\footnotemark[2] \\ \textbf{Longtao Huang}$^{1}$, \textbf{Hui Xue}$^{1}$, \textbf{Jun Huang}$^{1}$\\
$^1$ Alibaba Group
$^2$ East China Normal University\\
 {\tt \{zhangtaolin.ztl, chengyu.wcy\}@alibaba-inc.com, hexf@cs.ecnu.edu.cn} \\
 }

\begin{document}
\maketitle
\begin{abstract}
Recently, while large language models (LLMs) have demonstrated impressive results, they still suffer from hallucination, i.e., the generation of false information.
Model editing is the task of fixing factual mistakes in LLMs;
yet, most previous works treat it as a one-time task, paying little attention to ever-emerging mistakes generated by LLMs.
We address the task of sequential model editing (SME) that aims to rectify mistakes continuously.
A \textbf{D}ynamic \textbf{A}uxiliary \textbf{F}usion \textbf{Net}work (DAFNet) is designed to enhance the semantic interaction among the factual knowledge within the entire sequence, preventing catastrophic forgetting during the editing process of multiple knowledge triples.
Specifically, (1) for semantic fusion within a relation triple, we aggregate the intra-editing attention flow into auto-regressive self-attention with token-level granularity in LLMs. We further leverage multi-layer diagonal inter-editing attention flow to update the weighted representations of the entire sequence-level granularity.
(2) Considering that auxiliary parameters are required to store the knowledge for sequential editing, we construct a new dataset named \textbf{DAFSet}, fulfilling recent, popular, long-tail and robust properties to enhance the generality of sequential editing.
Experiments show DAFNet significantly outperforms strong baselines in single-turn and sequential editing.
The usage of DAFSet also consistently improves the performance of other auxiliary network-based methods in various scenarios \footnote{The code and pre-trained models will be available at \url{https://github.com/qizhou000/DAFNet}}.
    \end{abstract}

\begin{figure*}[!tb]
\centering
\includegraphics[width=15.5cm]{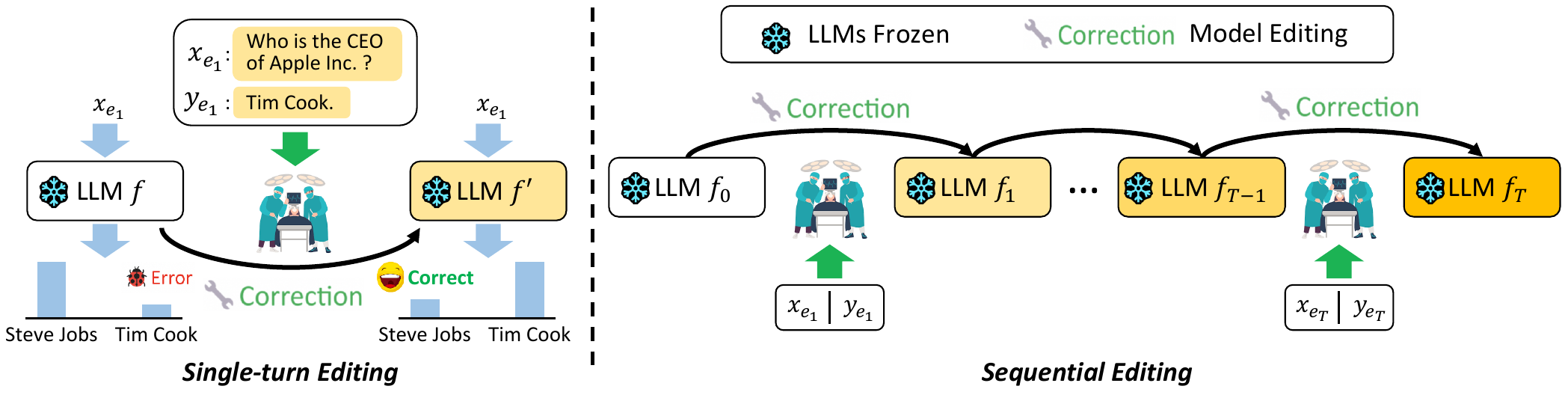}
\caption{The comparison between different model editing scenarios for LLMs including single-turn and sequential editing with $T$ steps. The single-turn editing model only edits one fact into the LLM at a time. In the sequential editing scenario, it requires to edit a series of facts continually (Best viewed in color).}
\label{motivation_example}
\end{figure*}

\section{Introduction}
Transformer-based models, particularly LLMs \cite{DBLP:conf/naacl/DevlinCLT19,DBLP:conf/nips/BrownMRSKDNSSAA20,DBLP:journals/corr/abs-2302-13971,DBLP:journals/fi/RoumeliotisT23} have become backbones of modern NLP and delivered promising results in various downstream tasks \cite{DBLP:conf/acl/LiGGC22,DBLP:conf/acl/ZhengZCHTL0M22,DBLP:conf/acl/BlevinsGZ23,DBLP:conf/acl/BlinovaZJEB23}.
However, LLMs still produce undesirable outputs occasionally \cite{DBLP:journals/nca/BastaCC21,DBLP:journals/corr/abs-2310-20689}.
The cost of such mistakes is non-negligible and exhibits an inclination to generate hallucinations \cite{ DBLP:conf/icml/ShiCMSDCSZ23,DBLP:conf/acl/TamMZKBR23}, resulting in seemingly plausible yet factually unsupported contents.
To alleviate these problems, there has been growing interest in integrating knowledge into LLMs through model editing \cite{DBLP:conf/emnlp/CaoAT21,DBLP:conf/emnlp/MadaanTCY22,DBLP:conf/iclr/MengSABB23}.
It updates the knowledge stored in relevant parameters in LLMs without fine-tuning the whole model.

In the literature, previous model editing methods\footnote{``Model editing'' and ``knowledge editing'' of LLMs share the same meaning. Thus, we use both terms interchangeably.} can be divided into two categories, including \texttt{Single-turn Editing} and \texttt{Sequential Editing}.
(1) \texttt{Single-turn Editing} edits one or a batch of knowledge triples at a time via modifying the original parameters of LLMs based on meta-learning \cite{DBLP:conf/emnlp/CaoAT21,DBLP:conf/iclr/MitchellLBFM22} and locate-then-edit methods \cite{DBLP:journals/corr/abs-2211-11031,DBLP:conf/nips/MengBAB22,DBLP:conf/iclr/MengSABB23}.
These works struggle to handle factual knowledge that is continuously updated in real-world scenarios.
(2) \texttt{Sequential Model Editing} (SME) based approaches learn the updated knowledge via adding extra modules \cite{DBLP:conf/emnlp/DongDSXSL22,DBLP:conf/icml/MitchellLBMF22,DBLP:conf/iclr/HuangSZZR023}.
The editing process of parameters is not achieved through back-propagation, but instead through weights independently calculated by extra modules associated with each fact.
However, these methods do not take into account the effects of mutual semantic influence between these sequentially input facts.
Another overlooked problem is that extra parameters are randomly initialized and can not provide a good starting point to cover the editing properties \cite{DBLP:conf/emnlp/YaoWT0LDC023} of the testing data.\footnote{The post-edit model should satisfy the following properties: reliability, generality and locality \cite{DBLP:conf/emnlp/YaoWT0LDC023}.}
As shown in Figure \ref{motivation_example}, previous model editing methods edit the sequential facts independently similar to single-turn editing in SME scenario.
Thus, subsequent LLMs need to wait for previous LLMs to finish editing before they can be edited.

In this paper, we propose a dynamic auxiliary fusion network named \textbf{DAFNet} to explicitly capture the correlation among the input sequential triples for the update of knowledgeable parameters in LLMs.
In addition, an auxiliary dataset named \textbf{DAFSet} for learning the meta-edit weights is constructed to compensate for the learning gap between extra cold-start parameters and testing properties in the training stage.
To our knowledge, this is currently the first comprehensive training set in model editing\footnote{Previous public datasets are used for testing editing ability of LLMs directly \cite{DBLP:conf/conll/LevySCZ17,DBLP:conf/nips/MengBAB22,DBLP:conf/iclr/MengSABB23}.}.
Specifically, we introduce the above two main contributions as follows:

\noindent\textbf{DAFNet:} Unlike previous methods that update parameters independently, we aggregate the representations of facts into intra- and inter-editing auto-regressive flows.
The intra-editing attention flow gathers each token-level fact representation auto-repressively via assigning different semantic weights to tokens, decomposing the LLMs' high-dimensional outputs to two low-rank representations as editing signal.
The inter-editing attention flow obtains interactive representations of sentence-level sequential inputs through multi-layer auto-regressive iterative diagonal attention between facts.
Finally, our designed loss based on the desired editing properties is leveraged to train the auxiliary network.

\noindent\textbf{DAFSet:} In previous works \cite{DBLP:conf/emnlp/CaoAT21,DBLP:conf/iclr/MitchellLBFM22,DBLP:journals/corr/abs-2311-04661},  weights of the original auxiliary network  are usually randomly initialized and then combined with LLMs in multi-task training.
Therefore, the lack of training set for existing editing methods causes the inconsistency with the distributional characteristics of the final editing target solely by editing the facts \cite{DBLP:conf/icml/BickelBS07,DBLP:journals/corr/abs-2303-02011}.
The goal of constructing our auxiliary dataset is to learn auxiliary meta-weights to compensate for the bias caused by random weight distributions.
DAFSet is designed to include four different properties (i.e., \texttt{Recency, Popularity, Long-tailness} and \texttt{Robustness}) based on the test editing properties.
We collect data for these properties via subject frequency and output likelihood in various domains to make the learned auxiliary weights equipped with more generalized editing abilities.



\section{Related Work}
In this section, we briefly overview the related works of model editing for LLMs in four aspects.

\noindent\textbf{Adding Extra Modules:} This method stores all edit examples in memory and uses a retriever to extract the most relevant facts for each new input, guiding the model in generating the edited fact.
SERAC \cite{DBLP:conf/icml/MitchellLBMF22} adopts a distinct counterfactual model while leaving the original model unchanged.
Other methods edit LLMs by prompting the model with the edited fact and retrieved edit demonstrations from the memory such as MemPrompt \cite{DBLP:conf/emnlp/MadaanTCY22}, IKE \cite{DBLP:journals/corr/abs-2305-12740} and MeLLo \cite{DBLP:conf/emnlp/ZhongWMPC23}.

\noindent\textbf{Additional Parameters:} This paradigm introduces additional trainable parameters in LLMs, which are trained using a modified knowledge dataset while keeping the original parameters unchanged.
T-Patcher \cite{DBLP:conf/iclr/HuangSZZR023} and CaliNET \cite{DBLP:conf/emnlp/DongDSXSL22} integrate a new single neuron (patch) for each error in the last layer of the FFN.
GRACE \cite{DBLP:journals/corr/abs-2211-11031} maintains a discrete codebook module for the middle layer of the LLMs.

\noindent\textbf{Locate-Then-Edit:} The initial step involves identifying parameters that correspond to specific knowledge and then modifying them through direct updates to the target parameters.
The work \cite{DBLP:conf/acl/DaiDHSCW22} introduces a method for identifying the specific ``knowledge neuron'' (a key-value pair in the FFN matrix) that represents the knowledge, and subsequently updating these neurons. 
ROME \cite{DBLP:conf/nips/MengBAB22} applies causal mediation analysis to locate the editing area.
MEMIT \cite{DBLP:conf/iclr/MengSABB23} expands on the setup of ROME, realizing the situation of synchronous editing for multiple cases.

\noindent\textbf{Meta-learning:} It utilizes a hyper-network to acquire the essential updated weights for editing.
Knowledge Editor (KE) \cite{DBLP:conf/emnlp/CaoAT21} utilizes a hyper-network to predict the weight updated for each data point. 
MEND \cite{DBLP:conf/iclr/MitchellLBFM22} learns to edit LLMs fastly
 improving the performance via employing a low-rank decomposition of gradients as the input.
MALMEN \cite{DBLP:journals/corr/abs-2311-04661} accommodates editing multiple facts with limited memory budgets via separating the computation on the hyper-network and LM enabling arbitrary batch size on both neural networks.
Note that, all the above editing methods focus on single-turn edits with one or a batch of facts, not taking into account semantic connections among the facts in a sequential order.

\begin{figure}[tb]
\centering
\includegraphics[width=\columnwidth]{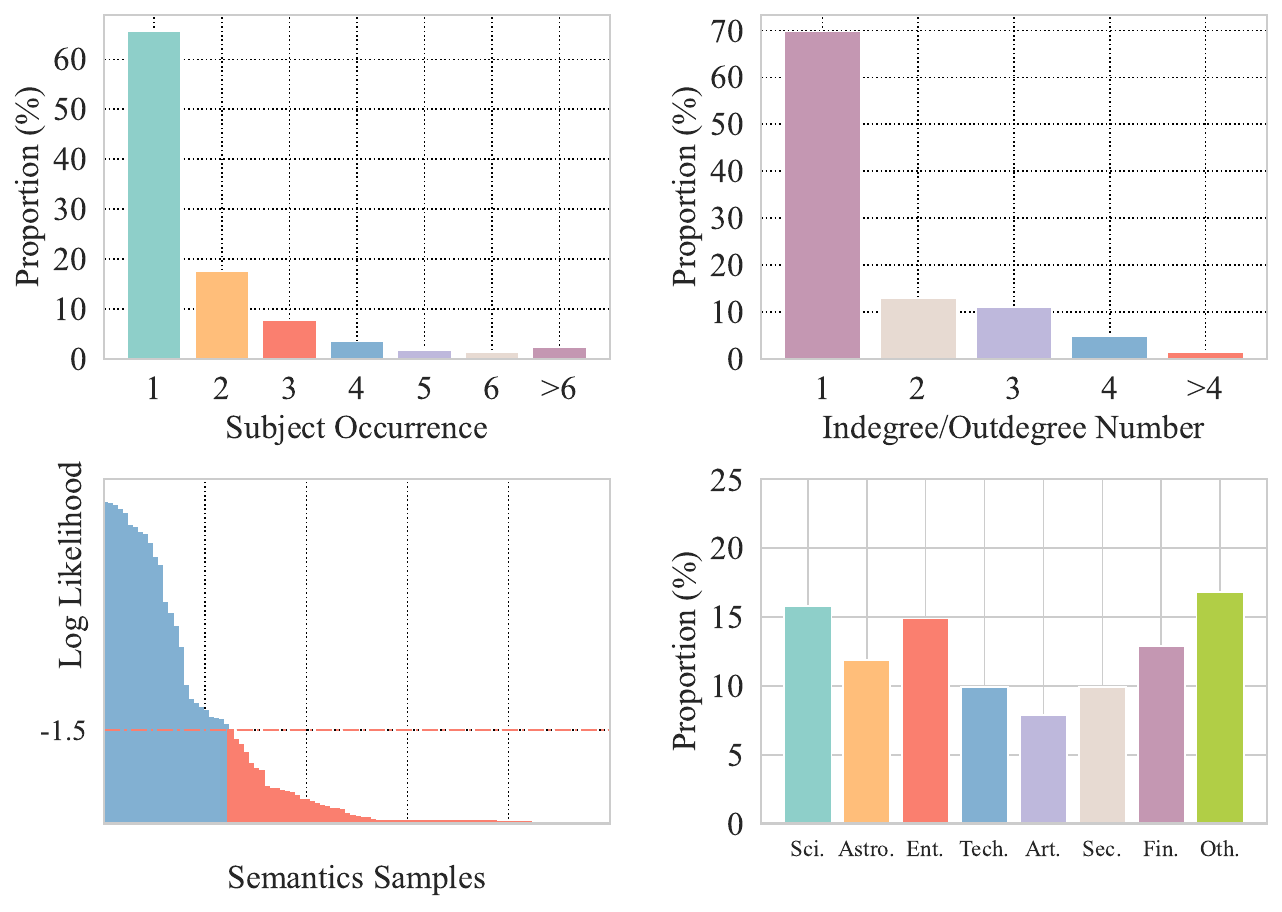}
\caption{Statistical results of the collected DAFSet.}
\label{long_tail}
\end{figure}

\section{Preliminaries of SME}
\label{define_sme}
In this section, we provide a brief introduction to SME for LLMs.
A model $f \in \mathbb{F}$ can be defined as a function $f: \mathbb{X} \mapsto \mathbb{Y}$ that maps an input $x$ to its prediction $f(x)$.
Then, given a model $f$ and an edit example pair $(x_e, y_e)$ that $f(x_e) \neq y_e$, a model editor (ME) outputs a post-edit model $f'$.
\begin{gather}
    \mathrm{ME}: \mathbb{F} \times \mathbb{X} \times \mathbb{Y} \mapsto \mathbb{F} \\
     f'=\mathrm{ME}(f,x_e,y_e)
\end{gather}
Given a sequence of facts ${(x_{e_1}, y_{e_1}),..., (x_{e_T}, y_{e_T})}$ and an initial model $f$, a model editor ME needs to conduct edits successively when the model makes undesirable output:
\begin{equation}
f_t= \operatorname{ME}\left(f_{t-1}, x_{e_t}, y_{e_t}\right),\, t=1,...,T 
\end{equation}
where assume $f_0=f$.
Every edit in SME should satisfy the following three properties:

\paragraph{Reliability} A reliable edit holds when the post-edit model $f_t$ gives the target answer for the every cases $(x_{e_\tau}, y_{e_\tau}),\tau\leq t$ to be edited. The reliability is measured as the average accuracy on the edit cases:
\begin{equation}
        \mathbb{E}_{\left(x_e, y_e\right) \sim\left\{\left(x_{e_\tau}, y_{e_\tau}\right)\right\}_{\tau=1}^t} \mathbb{I}\left\{\mathop{\operatorname{argmax}}\limits_y f_t\left(y \mid x_e\right)=y_e\right\}
\end{equation}
\paragraph{Generality}  The post-edit model $f_t$ should also satisfy the relevant neighbours $N(x_{e_\tau}, y_{e_\tau}), \tau\leq t$.
It is evaluated by the average accuracy of $f_t$ on examples drawn uniformly from the relevant neighborhood:
\begin{equation}
\begin{split}
    \mathbb{E}_{\left(x_e, y_e\right) \sim\left\{\left(x_{e_\tau}, y_{e_\tau}\right)\right\}_{\tau=1}^t}\mathbb{E}_{(x_g, y_g) \sim N\left(x_{e}, y_{e}\right)} \operatorname{G}(x_g, y_g)\\
    \text{s.t.}\;\operatorname{G}(x_g, y_g) = \mathbb{I}\left\{\mathop{\operatorname{argmax}}\limits_y f_t \left(y \mid x_g\right)=y_g\right\}
\end{split}
\end{equation}
\paragraph{Locality} Editing should be implemented locally, which means the post-edit model $f_t$ should not change the output of irrelevant examples in out-of-scope $O(x_{e_\tau}, y_{e_\tau}), \tau\leq t$. Hence, the locality is evaluated by the rate at which the post-edit model $f_t$’s predictions are unchanged as the pre-edit model $f$:
\begin{equation}
\begin{split}
    \mathbb{E}_{\left(x_e, y_e\right) \sim\left\{\left(x_{e_\tau}, y_{e_\tau}\right)\right\}_{\tau=1}^t} \mathbb{E}_{(x_l, y_l) \sim O(x_{e}, y_{e})} \operatorname{L}(x_l, y_l)\\
    \text{s.t.}\;\operatorname{L}(x_l, y_l) = \mathbb{I}\left\{f_t \left(y \mid x_l\right)=f\left(y \mid x_l\right)\right\}
\end{split}
\end{equation}

\section{The DAFSet Dataset}
Due to the lack of training editing sets for various meta-learning based methods,
we propose DAFSet to provide the initial editing ability of LLMs.
It enables the model to better emerge with knowledge generalization ability \cite{DBLP:conf/emnlp/YaoWT0LDC023,DBLP:journals/corr/abs-2307-12976} in the test editing stage with four properties.

\begin{figure*}[!tb]
\centering
\includegraphics[width=15.5cm]{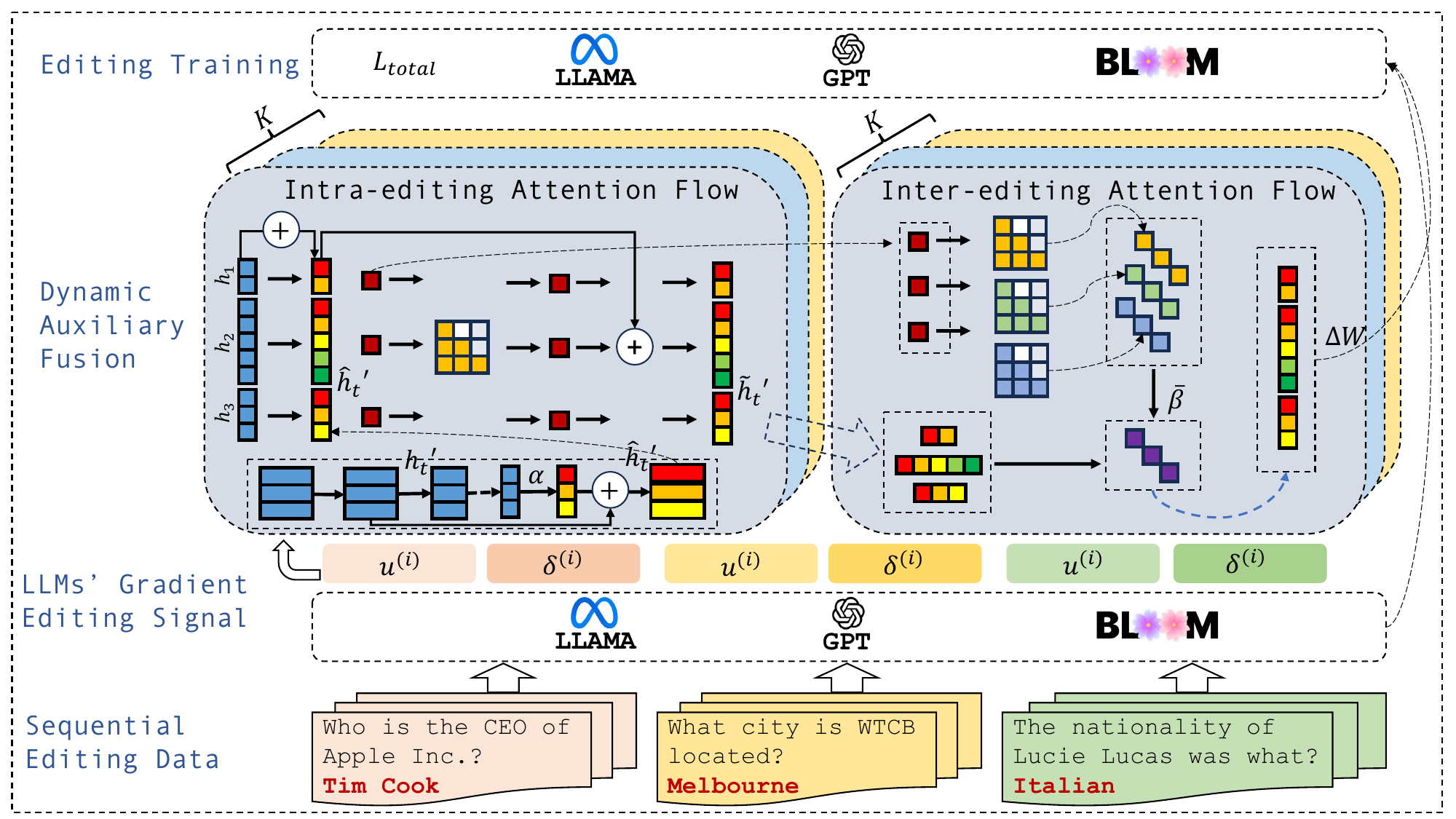}
\caption{Model overview. Our DAFNet model mainly includes three steps: \texttt{Gradient Editing Signal Acquisition}, \texttt{Dynamic Auxiliary Fusion} and \texttt{Editing Training}. Particularly, \texttt{Dynamic Auxiliary Fusion} is designed with intra and inter-editing attention flows to capture the interaction between editing facts.}
\label{model_main}
\end{figure*}

\subsection{Data Collection}
We leverage \texttt{Wikidata}\footnote{\url{https://en.wikipedia.org/wiki/Wikidata:Main\_Page}}, a knowledge graph (KG) with relation triples represented as $(e_h, r, e_t)$, where $e_h$ is the head entity, $r$ is the relation predicate and $e_t$ is the tail entity.
The specific collection steps for these properties are as follows.


\noindent\textbf{Recency:} We gather triplets that have been recently added to Wikidata.
The head $e_h$ and tail entities $e_t$ are often associated with numerous redundant triples. To address this, we only collect relation triples associated with a set of 48 common relation predicates. 
We use the off-the-shelf tool\footnote{\url{https://pypi.org/project/qwikidata/0.2.0/}} to search the triples where each triple has been modified in the last 7 days. Then, we use templates to map the triples to construct the data samples, with templates shown in Appendix \ref{appendix_dafset}.

\noindent\textbf{Popularity:} 
We collect triples corresponding to popular entities, where head entities are from top-viewed pages in Wikipedia.
Next, we perform multi-hop selection of tail entities within 2-hops.
Finally, we also leverage the above templates to construct the training samples.

\noindent\textbf{Long-tailness:} LLMs often lack sufficient learning of low-frequency data, and thus editing effect on such knowledge is poor. We identify and construct training data related to long-tail knowledge 
from three perspectives: (1) \texttt{Frequency}: we first collect head entities 
and count their corresponding frequencies in Wikidata. Then, we set a threshold to select approximately 80\% of entities with low frequencies. (2) \texttt{In-out KG Degree}: we calculate the connectivity of the corresponding head entities in KG, and then set the threshold to select low-frequency entities. (3) \texttt{Likelihood}: unlike the two intuitive statistics, our approach to identify long-tail data focuses on the semantics of the model's output. 
Specifically, we input the sentence related to the head entity into LLMs and evaluate the model's comprehension of the entity by examining the likelihood probability of its position\footnote{Detailed construction process is shown in Appendix \ref{appendix_dafset}.}.

\noindent\textbf{Robustness:} We employ three robustness properties \cite{DBLP:journals/access/OmarCNM22} for constructing the training data, including text length, context, and emotions. To control the length of input prompts, we utilize the ``loc'' and ``rephrase'' fields of each edited data \cite{DBLP:conf/conll/LevySCZ17,DBLP:conf/iclr/MitchellLBFM22,DBLP:journals/corr/abs-2307-12976}. Templates are used for training sentence construction, specifically for context and emotions, to prompt the LLMs for generation. To further enhance the robustness, we generate two opposite data attributes for each training data point,  such as ``positive'' and ``negative'' in emotions.

To learn auxiliary parameters and provide an initial meta-weight for editing, 
the data labeled as ``recent'' and ``popular'' is predominantly stable and frequently used to insert factual information. This leads to improved reliability and locality of test editing.
In terms of the generality metric for test editing, the ``long-tailness'' and ``robustness'' categories can enhance the meta-weights, considering the sparsity of editing data.
Overall, there are a total of 30,000 final constructed data samples for long-tailness, 13,000 for robustness, 30,000 for popularity, and 30,000 for recency.

\subsection{Statistical Analysis on Our Dataset}
We further conduct analysis on two complicated properties, i.e., 
 long-tailness and robustness.
As shown in Figure \ref{long_tail}, we visualize the statistical results of our training data.
The first subgraph on the left shows the frequency of head entities. We can see that the lower the frequency, the greater the proportion such as the sum of occurrence 1 and 2 has exceeded 80\%.
The second subgraph shows the proportion of head entities with different in-out kG' degrees.
We see that most of the data has neighboring edges with high degrees and thus we select long-tail triples with low in-out degrees.
The third subgraph shows the output likelihood probability of target entities in editing sentences.
We select data samples with low log likelihood (i.e., $<-1.5$) output corresponding to a large proportion as our long-tail training sentences.
In the fourth subgraph, given that robustness needs to be applicable across various domains, we analyze the distribution of domains in the generated training data for robust editing. 

\section{The DAFNet Model}
In this section, we formally introduce our DAFNet model, with an overall architecture shown in Figure~\ref{model_main}.

In DAFNet, to obtain rich semantic representation for each knowledge triple, we perform a token-level granularity fusion modeling by intra-editing attention flow in Section \ref{sub_intra_editing}.
Next, we leverage the auto-regressive modeling process to naturally capture the semantic interaction among sequential facts by inter-editing attention flow in Section \ref{subsection_inter_attention_flow}.

\subsection{Editing Signal Acquisition}
In auxiliary network-based methods, we need to first determine the editing position before performing editing.
We heuristically select linear layers at the last few layers of LLMs \cite{DBLP:conf/nips/MengBAB22,DBLP:conf/iclr/MitchellLBFM22}.
Then, we obtain the editing signal input to the auxiliary network by decomposing gradients of the editing weight matrix.

Specifically, the gradient of the editing loss w.r.t.
the weight matrix (to be edited) of a certain linear layer $W$ in a language model $f$ is a summation of $B$ rank-1 matrix, where $B$ is the token count of an editing sample $(x_e,y_e)$. Formally, we have
\begin{gather}
\begin{split}
\nabla_{W} \mathcal{L}_{edit}(x_e,y_{e})=\sum_{i=1}^B u^{(i) \top}\cdot\delta^{(i)} \\
\text{s.t.}\; \mathcal{L}_{edit}(x_e,y_{e}) = -\operatorname{log} f(y_e|x_e)
\end{split}
\end{gather}
where $u^{(i)}\in \mathbb{R}^{1 \times d_{in}}, \delta^{(i)}\in \mathbb{R}^{1 \times d_{out}}$ are the input and
the output gradient of the linear layer at the $i_{th}$ token position, respectively. 
Thus, the editing signal of weight $W$ for editing sample $(x_e,y_e)$ is defined as $h = \left[u;\delta\right]\in\mathbb{R}^{B\times (d_{in}+d_{out})}$.
Note that editing signals w.r.t. different $W$s are independently fed into auxiliary networks.


\subsection{Intra-editing Attention Flow}
\label{sub_intra_editing}
In SME, the semantic modeling among facts has a significant influence on the overall performance. 
We first separately fuse the edited signal after dimensional reduction w.r.t. each token, and then aggregate the fused edited sequence fact into auto-regressive self-attention to further enhance the interaction between fact tokens.

Let the editing signal of the $t_{th}$ editing fact as $h_t^0 \in \mathbb{R}^{B_t \times (d_{in} + d_{out})}$, which is the input to the first intra-editing attention module. 
Assume the input to the $k_{th}$ layer is $h_t^{k-1}$. 
For the sake of simplicity, the layer symbol $k$ is omitted below in this subsection.
We obtain the token attention score $\alpha_i$ via allocating the representation importance.
The aggregated fact representations $\widehat{h}_{t} \in \mathbb{R}^{d_{in}+d_{out}}$ fuse the tokens' representations together with the token attention score.
Specifically, we first transform the editing signal output into fused representation $h_{t}' \in \mathbb{R}^{B \times (d_{in}+d_{out})}$ through a residual module as:
    \begin{equation}
h_t' = (\sigma (  h_{t} W_1 + b_1)) W_2   + b_2 
\end{equation}
where $W_1 \in \mathbb{R}^{(d_{in} + d_{out}) \times d_{down}}$ and $W_2 \in \mathbb{R}^{d_{down} \times (d_{in} + d_{out})}$ are the linear layer weights.
$\sigma$ is ReLU function \cite{agarap2018deep}.
Then, we calculate the token attention weight $\alpha \in \mathbb{R}^{ B_t \times 1}$ using the transformed representation $h_{t}^{\prime}$. 
The fused editing representation $\widehat{h}_{t}$ is learned from the token-span representation $h_{t}^{\prime}$ as follows:
\begin{gather}
    \alpha = \varphi( \sigma( h_{t}' W_3  +b_3) W_4  + b_4), \widehat{h}_{t} = \alpha \otimes h_{t}^{\prime} \\
\Bar{h}_{t} = \sum\limits_{i=1}^{B_t} \widehat{h}_{t}^{(i)}, \quad \widehat{h}_{t}' = \widehat{h}_{t} + h_{t}
\end{gather}
where $W_{3} \in \mathbb{R}^{(d_{in} + d_{out}) \times d_{down}}$, $W_{4} \in \mathbb{R}^{d_{down} \times 1}$.
$\varphi$ is the softmax non-linear activation function.
$\otimes$ is the element-wise multiplication.
Finally, each fused fact representation $\Bar{h}_{t} \in \mathbb{R}^{1 \times (d_{in} + d_{out})}$ is aggregated by the token importance.
After obtaining the fused representations of all facts in the sequence, we transform the fused representations into the auto-regressive model for iteration learning to enhance the semantic information modeling between the fused facts:
\begin{gather}
    \Bar{h}_{1}', \Bar{h}_{2}', ..., \Bar{h}_{T}' = f_{intra}(\Bar{h}_{1}, \Bar{h}_{2}, ..., \Bar{h}_{T}) \\ 
    \tilde{h}_{t} = \widehat{h}_{t}' \oplus \Bar{h}_{t}', \quad t= 1,2,...,T
\end{gather}
where $f_{intra}$ is the auto-regressive self-attention layer to fuse the facts sequentially.
$\Bar{h}_{t}' \in \mathbb{R}^{1 \times (d_{in} + d_{out})}$ is the $i_{th}$ auto-regressive fact representation. $T$ is the number of facts for sequential editing modeling. $\oplus$ is the element-wise addition.
$\tilde{h}_{t} \in \mathbb{R}^{B_t \times (d_{in}+d_{out})}$ is the final intra-editing attention fusion representation. 
The intra-editing attention flow module is stacked by $K$ layers. 
Finally, $\tilde{h}_{t}$ is fed as the input into the next module.

\subsection{Inter-editing Attention Flow}
\label{subsection_inter_attention_flow}
Considering the effect of enhancing the interaction between facts in a sequence on updating iteration weights, we use the multi-layer auto-regressive diagonal attention to fuse the previous edited fact representations within a sequence.
For each achieved editing fact weight score, we aggregate the last layer editing fusion weight representation $\tilde{h}_{t},t=1,...,T$ to update the original LLMs.

Specifically, we use another multi-layer auto-regressive self-attention layer to learn the relationship of facts for sequential editing modeling:
\begin{equation}
\beta^k=\operatorname{diag}\left(f_{sa}^k\left(\Bar{h}_{1}, \Bar{h}_{2}, \ldots, \Bar{h}_{T}\right)\right), k=1,...,K 
\end{equation}
where $f_{sa}^k$ is the function to obtain the auto-regressive self-attention matrix at $k_{th}$ layer.
$\operatorname{diag}(*)$ is the function to achieve diagonal values of a matrix.
$\beta^k \in \mathbb{R}^{T}$.
Then, we aggregate the diagonal results of $K$ layers into the averaged facts' importance:
$\Bar{\beta}=\frac{1}{K}\sum_{k=1}^K \beta^k$
where $\Bar{\beta} \in \mathbb{R}^{T}$.
Finally, the inter-editing fusion representations of each fact 
is $\tilde{h}_{t}^K \in \mathbb{R}^{B_t \times (d_{in}+d_{out})}$ at last $K_{th}$ layer.
We recover it
to the original LLM's weight shape to obtain updated weights of the $T$ sequential facts and then sum them weighted by $\Bar{\beta}$:
    \begin{gather}
    [\tilde{u}_{t}; \tilde{\delta}_{t}] = \tilde{h}_{t}^K, \ \ \   \Delta W_t = \frac{\tilde{u}_{t}^\top \otimes \tilde{\delta}_{t}}{B_t} \\
    \Delta \tilde{W}_T =\sum_{t=1}^T\left(\prod_{\tau=t+1}^{T}1-\Bar{\beta}_\tau\right)
    \Bar{\beta}_t \cdot\Delta W_t \label{formula_updated_weights_summation}
\end{gather}
where $\tilde{u}_{t} \in \mathbb{R}^{B \times d_{in}}$ and $\tilde{\delta}_{t} \in \mathbb{R}^{B\times d_{out}}$ are the decomposed weight representations, respectively.
$\Delta \tilde{W}_T \in \mathbb{R}^{d_{in} \times d_{out}}$ aggregates all the token weight representations together to achieve the fact updated weights' representation.
Finally, the process of sequential editing by $T$ facts can be formulated as:
$f_T=\operatorname{\Gamma}(f_{T-1},\Delta \tilde{W}_T)$
where $\operatorname{\Gamma}$ indicates adding $\Delta\tilde{W}_T$ to the corresponding matrix to be edited. Algorithm \ref{alg_dafnet_edit_once} in the appendix describes the implementation of how DAFNet performs each edit one by one in sequential editing. 
Next we formulate the losses of $f_T$ to model $T$ edits into DAFNet.

\subsection{Sequential Editing Training}
Our auxiliary network considers three editing properties including reliability, generality and locality.
Hence, the total loss with a $T$ sequential editing facts is defined as follows:
\begin{gather}
    \mathcal{L}_{rel}(f_T) = \sum_{t=1}^T -\log{f_T(y_e^{(t)}|x_e^{(t)})} \\ 
    \mathcal{L}_{gen}(f_T) = \sum_{t=1}^T\sum_{j=1}^{N_{g}^{(t)}} -\log{f_T(y_{g_j}^{(t)}|x_{g_j}^{(t)})} \\
    \mathcal{L}_{loc}(f,f_T) = \sum_{t=1}^T \sum_{j = 1}^{N_{l}^{(t)}}\mathrm{KL}(f(x_{l_j}^{(t)})||f_T(x_{l_j}^{(t)})) \\
    \mathcal{L}_{total} = \mathcal{L}_{rel}(f_T) + \mathcal{L}_{gen}(f_T) + \mathcal{L}_{loc}(f,f_T)
\end{gather}
where $(x_e^{(t)}, y_e^{(t)})$ is the reliability sample of the $t_{th}$ fact, i.e., the editing sample itself.
$(x_{g_j}^{(t)}, y_{g_j}^{(t)})$ and 
$x_{l_j}^{(t)}$ 
are the $j_{th}$ generality and locality sample of $t_{th}$ fact, respectively.
$N_{g}^{(t)}$ and $N_{l}^{(t)}$ are the corresponding loss sample number of $t_{th}$ fact.
$\mathrm{KL}$ is the Kullback-Leibler Divergence function. 
Algorithm \ref{alg_dafnet_train} describes the training process of DAFNet.

\begin{table*}[!tb]
\scriptsize
\centering
\setlength{\tabcolsep}{4.3pt}
\renewcommand{\arraystretch}{0.85}
\begin{tabular}{ccccccccccccccc}
\midrule
\multirow{2}{*}{\textbf{Backbone}} & \multirow{2}{*}{\textbf{\# Editing}} & \multirow{2}{*}{\textbf{Editor}} & \multicolumn{4}{c}{\textbf{ZSRE}}                 & \multicolumn{4}{c}{\textbf{CounterFact}}                   & \multicolumn{4}{c}{\textbf{RIPE}}                 \\
                                &                               &                                  & \textbf{Rel.} & \textbf{Gen.} & \textbf{Loc.} & \textbf{Avg.} & \textbf{Rel.} & \textbf{Gen.} & \textbf{Loc.} & \textbf{Avg.} & \textbf{Rel.} & \textbf{Gen.} & \textbf{Loc.} & \textbf{Avg.} \\ \midrule

\multirow{24}{*}{\makecell[c]{GPT-J \\ (6B)} }

&\multirow{10}{*}{10}&FT&10.3&10.8&0.3&7.1$_{(\pm0.1)}$&56.2&24.2&2.1&27.5$_{(\pm0.5)}$&7.8&4.3&1.4&4.5$_{(\pm0.1)}$\\
&&TP&85.2&78.3&77.2&80.2$_{(\pm1.2)}$&96.0&54.3&3.6&51.3$_{(\pm1.2)}$&80.8&56.7&32.4&56.6$_{(\pm1.7)}$\\
&&KN&1.0&1.1&1.9&1.3$_{(\pm0.0)}$&1.2&0.7&2.3&1.4$_{(\pm0.0)}$&0.1&0.3&0.2&0.2$_{(\pm0.0)}$\\
&&ROME&81.1&78.8&94.6&84.8$_{(\pm1.7)}$&95.9&59.4&90.0&81.8$_{(\pm1.9)}$&98.2&41.9&39.1&59.7$_{(\pm0.8)}$\\
&&MEMIT&82.1&76.0&94.7&84.2$_{(\pm2.0)}$&96.0&38.1&\textbf{95.5}&76.5$_{(\pm2.5)}$&98.5&37.7&47.3&61.2$_{(\pm1.2)}$\\
&&GRACE&81.8&78.4&94.5&84.9$_{(\pm1.6)}$&95.2&60.3&91.2&82.2$_{(\pm1.6)}$&98.0&40.9&38.7&59.2$_{(\pm0.4)}$\\
&&$\text{KE}^\spadesuit$&0.0&0.0&0.7&0.3$_{(\pm0.0)}$&0.0&0.0&0.2&0.1$_{(\pm0.0)}$&0.0&0.0&0.1&0.0$_{(\pm0.0)}$\\
&&$\text{MEND}^\spadesuit$&0.4&0.4&0.5&0.4$_{(\pm0.0)}$&0.6&0.2&0.2&0.3$_{(\pm0.0)}$&0.0&0.0&0.0&0.0$_{(\pm0.0)}$\\
&&$\text{MALMEN}^\spadesuit$&99.1&95.3&92.8&95.8$_{(\pm1.6)}$&90.0&32.9&77.1&66.7$_{(\pm2.2)}$&89.7&52.1&51.3&64.4$_{(\pm1.8)}$\\
&&$\text{DAFNet}^\spadesuit$&\textbf{99.6}&\textbf{97.6}&\textbf{94.8}&\textbf{97.3}$_{(\pm1.5)}$&\textbf{96.2}&\textbf{65.8}&85.2&\textbf{82.4}$_{(\pm1.6)}$&\textbf{98.7}&\textbf{57.6}&\textbf{57.6}&\textbf{71.3}$_{(\pm1.6)}$\\

\cmidrule{2-15}
&\multirow{10}{*}{100}&FT&2.2&1.9&0.3&1.4$_{(\pm0.0)}$&35.9&10.8&1.6&16.1$_{(\pm0.3)}$&5.7&1.6&0.1&2.5$_{(\pm0.1)}$\\
&&TP&68.5&59.3&52.8&60.2$_{(\pm1.3)}$&76.0&31.9&2.2&36.7$_{(\pm0.8)}$&64.2&36.4&23.7&41.4$_{(\pm1.0)}$\\
&&KN&0.6&0.4&0.8&0.6$_{(\pm0.0)}$&0.2&0.5&0.8&0.5$_{(\pm0.0)}$&0.0&0.0&0.0&0.0$_{(\pm0.0)}$\\
&&ROME&77.4&75.6&85.0&79.3$_{(\pm2.2)}$&78.8&38.4&52.2&56.5$_{(\pm1.0)}$&\textbf{95.7}&36.0&32.2&54.6$_{(\pm1.0)}$\\
&&MEMIT&77.9&74.1&90.2&80.7$_{(\pm2.5)}$&\textbf{94.1}&40.2&85.1&\textbf{73.1}$_{(\pm1.3)}$&86.6&33.3&33.5&51.1$_{(\pm1.3)}$\\
&&GRACE&77.8&74.6&85.9&79.4$_{(\pm2.0)}$&76.3&39.2&51.6&55.7$_{(\pm0.8)}$&94.8&36.7&31.5&54.3$_{(\pm0.8)}$\\
&&$\text{KE}^\spadesuit$&0.0&0.0&0.7&0.2$_{(\pm0.0)}$&0.0&0.0&0.1&0.0$_{(\pm0.0)}$&0.0&0.0&0.0&0.0$_{(\pm0.0)}$\\
&&$\text{MEND}^\spadesuit$&0.2&0.1&0.0&0.1$_{(\pm0.0)}$&0.2&0.2&0.0&0.1$_{(\pm0.0)}$&0.0&0.0&0.1&0.0$_{(\pm0.0)}$\\
&&$\text{MALMEN}^\spadesuit$&50.6&40.7&59.3&50.2$_{(\pm0.8)}$&29.7&31.8&68.0&43.2$_{(\pm0.4)}$&39.9&27.8&53.2&40.3$_{(\pm0.8)}$\\
&&$\text{DAFNet}^\spadesuit$&\textbf{89.5}&\textbf{76.5}&\textbf{90.2}&\textbf{85.4}$_{(\pm1.6)}$&81.8&\textbf{40.3}&\textbf{87.3}&69.8$_{(\pm1.5)}$&78.5&\textbf{38.9}&\textbf{64.4}&\textbf{60.6}$_{(\pm1.5)}$\\

\midrule

\multirow{24}{*}{\makecell[c]{LLAMA2 \\ (7B)}}

&\multirow{10}{*}{10}&FT&38.3&37.4&57.9&44.5$_{(\pm0.9)}$&19.3&13.6&22.3&18.4$_{(\pm0.2)}$&30.5&21.8&28.0&26.8$_{(\pm0.7)}$\\
&&TP&57.3&52.4&36.7&48.8$_{(\pm1.1)}$&85.9&58.6&21.5&55.4$_{(\pm0.9)}$&63.4&41.2&30.4&45.0$_{(\pm0.8)}$\\
&&KN&0.0&0.0&0.7&0.2$_{(\pm0.0)}$&0.8&0.7&4.4&1.9$_{(\pm0.1)}$&0.0&0.0&0.3&0.1$_{(\pm0.0)}$\\
&&ROME&41.1&39.6&93.0&57.9$_{(\pm1.4)}$&38.6&24.9&83.6&49.1$_{(\pm1.0)}$&33.4&20.3&29.5&27.7$_{(\pm0.6)}$\\
&&MEMIT&24.3&24.1&51.1&33.2$_{(\pm0.9)}$&18.6&15.4&62.9&32.3$_{(\pm0.7)}$&18.4&13.6&10.1&14.1$_{(\pm0.3)}$\\
&&GRACE&42.5&39.7&92.5&58.2$_{(\pm1.5)}$&38.1&24.5&82.6&48.4$_{(\pm0.8)}$&31.4&20.8&29.1&27.1$_{(\pm0.5)}$\\

&&$\text{KE}^\spadesuit$&0.5&0.5&1.4&0.8$_{(\pm0.0)}$&0.0&0.0&0.2&0.1$_{(\pm0.0)}$&0.1&0.2&1.2&0.5$_{(\pm0.0)}$\\
&&$\text{MEND}^\spadesuit$&0.3&0.3&3.3&1.3$_{(\pm0.0)}$&0.0&0.0&0.1&0.0$_{(\pm0.0)}$&0.3&0.2&1.6&0.7$_{(\pm0.0)}$\\
&&$\text{MALMEN}^\spadesuit$&96.2&88.3&92.6&92.4$_{(\pm1.8)}$&79.5&45.8&36.2&53.8$_{(\pm1.1)}$&84.7&47.5&70.9&67.7$_{(\pm2.3)}$\\
&&$\text{DAFNet}^\spadesuit$&\textbf{97.2}&\textbf{92.0}&\textbf{93.3}&\textbf{94.1}$_{(\pm1.2)}$&\textbf{87.3}&\textbf{59.6}&\textbf{85.9}&\textbf{77.6}$_{(\pm1.4)}$&\textbf{88.8}&\textbf{56.4}&\textbf{83.2}&\textbf{76.1}$_{(\pm1.9)}$\\

\cmidrule{2-15}
&\multirow{10}{*}{100}&FT&7.6&7.0&4.1&6.3$_{(\pm0.2)}$&1.0&0.2&3.6&1.6$_{(\pm0.0)}$&1.8&0.8&1.0&1.2$_{(\pm0.1)}$\\
&&TP&46.1&41.2&9.7&32.3$_{(\pm0.5)}$&70.0&40.8&4.5&38.4$_{(\pm0.8)}$&44.7&28.9&11.6&28.4$_{(\pm0.7)}$\\
&&KN&0.0&0.0&0.0&0.0$_{(\pm0.0)}$&0.0&0.0&0.0&0.0$_{(\pm0.0)}$&0.0&0.0&0.0&0.0$_{(\pm0.0)}$\\
&&ROME&9.6&10.5&22.0&14.0$_{(\pm0.4)}$&33.6&22.1&68.0&41.2$_{(\pm1.2)}$&5.9&4.2&5.2&5.1$_{(\pm0.1)}$\\
&&MEMIT&0.7&0.7&0.9&0.8$_{(\pm0.0)}$&0.6&0.6&3.7&1.6$_{(\pm0.0)}$&0.2&0.5&0.3&0.3$_{(\pm0.0)}$\\
&&GRACE&9.3&8.5&23.0&13.6$_{(\pm0.5)}$&31.6&21.1&69.0&40.6$_{(\pm1.0)}$&5.7&4.9&5.1&5.2$_{(\pm0.2)}$\\
&&$\text{KE}^\spadesuit$&0.0&0.0&0.1&0.0$_{(\pm0.0)}$&0.0&0.0&0.9&0.3$_{(\pm0.0)}$&0.1&0.1&0.0&0.1$_{(\pm0.0)}$\\
&&$\text{MEND}^\spadesuit$&0.0&0.0&0.1&0.0$_{(\pm0.0)}$&0.0&0.0&0.1&0.0$_{(\pm0.0)}$&0.0&0.0&0.0&0.0$_{(\pm0.0)}$\\
&&$\text{MALMEN}^\spadesuit$&54.3&51.8&65.3&57.1$_{(\pm0.9)}$&48.1&22.4&47.2&39.2$_{(\pm0.6)}$&41.6&31.7&38.5&37.3$_{(\pm0.7)}$\\
&&$\text{DAFNet}^\spadesuit$&\textbf{84.7}&\textbf{72.0}&\textbf{93.6}&\textbf{83.5}$_{(\pm1.4)}$&\textbf{72.8}&\textbf{41.5}&\textbf{76.4}&\textbf{63.6}$_{(\pm1.1)}$&\textbf{57.7}&\textbf{41.2}&\textbf{87.5}&\textbf{62.2}$_{(\pm1.7)}$\\

\midrule
                                
\end{tabular}
\caption{The overall results of DAFNet and baselines in sequential edits. ``\# Editing'' indicates the length of sequential editing. ``Rel.'', ``Gen.'' and ``Loc.'' are the reliable, generality and locality editing metrics, respectively. The t-tests demonstrate the improvements of DAFNet are statistically significant with $p$ < 0.05 level. The editors marked with $\spadesuit$ are methods requiring training before editing, which are all augmented with DAFSet in this table.}
\label{main_exp}
\end{table*}

\begin{table*}[!tb]
\scriptsize
\centering
\setlength{\tabcolsep}{4.3pt}
\renewcommand{\arraystretch}{0.85}
\begin{tabular}{ccccccccccccccc}
\midrule
\multirow{2}{*}{\textbf{Backbone}} & \multirow{2}{*}{\textbf{\# Editing}} & \multirow{2}{*}{\textbf{Editor}} & \multicolumn{4}{c}{\textbf{ZSRE}}                 & \multicolumn{4}{c}{\textbf{CounterFact}}                   & \multicolumn{4}{c}{\textbf{RIPE}}                 \\
                                &                               &                                  & \textbf{Rel.} & \textbf{Gen.} & \textbf{Loc.} & \textbf{Avg.} & \textbf{Rel.} & \textbf{Gen.} & \textbf{Loc.} & \textbf{Avg.} & \textbf{Rel.} & \textbf{Gen.} & \textbf{Loc.} & \textbf{Avg.} \\ \midrule


\multirow{10}{*}{\makecell[c]{LLAMA2 \\ (7B)}}

&\multirow{10}{*}{1}&FT&52.8&53.2&92.1&66.0$_{(\pm0.3)}$&34.1&25.4&49.8&36.4$_{(\pm0.5)}$&49.2&33.7&71.0&51.3$_{(\pm0.5)}$\\
&&TP&86.4&84.0&86.4&85.6$_{(\pm1.4)}$&91.4&68.6&39.0&66.3$_{(\pm0.3)}$&77.0&55.1&51.3&61.1$_{(\pm0.5)}$\\
&&KN&20.2&20.8&52.4&31.1$_{(\pm0.1)}$&12.3&9.2&67.9&29.8$_{(\pm0.4)}$&21.8&15.4&55.4&30.9$_{(\pm0.5)}$\\
&&ROME&53.5&51.6&94.0&66.4$_{(\pm0.7)}$&41.1&21.8&91.8&51.6$_{(\pm0.8)}$&48.3&27.1&42.5&39.3$_{(\pm0.9)}$\\
&&MEMIT&49.7&49.4&91.9&63.7$_{(\pm0.5)}$&45.4&29.3&92.9&55.9$_{(\pm0.4)}$&58.4&29.6&38.7&42.2$_{(\pm0.3)}$\\
&&GRACE&52.3&50.8&95.7&66.3$_{(\pm0.9)}$&44.6&28.5&93.4&55.5$_{(\pm0.5)}$&56.7&30.6&41.1&42.8$_{(\pm0.2)}$\\

&&$\text{KE}^\spadesuit$&12.9&8.6&90.6&37.4$_{(\pm0.2)}$&8.0&2.9&90.3&33.7$_{(\pm0.4)}$&9.9&4.4&42.8&19.0$_{(\pm0.2)}$\\
&&$\text{MEND}^\spadesuit$&73.8&70.3&66.1&70.1$_{(\pm0.5)}$&81.1&67.2&77.1&75.1$_{(\pm0.3)}$&66.4&29.4&29.7&41.8$_{(\pm0.5)}$\\
&&$\text{MALMEN}^\spadesuit$&66.4&67.8&43.7&59.3$_{(\pm0.5)}$&52.4&42.3&36.6&43.8$_{(\pm0.6)}$&51.5&33.8&20.5&35.3$_{(\pm0.9)}$\\
&&$\text{DAFNet}^\spadesuit$&\textbf{97.5}&\textbf{97.4}&\textbf{94.9}&\textbf{96.6}$_{(\pm1.5)}$&\textbf{92.0}&\textbf{86.7}&\textbf{94.3}&\textbf{91.0}$_{(\pm0.8)}$&\textbf{97.8}&\textbf{66.4}&\textbf{72.2}&\textbf{78.8}$_{(\pm0.7)}$\\ \midrule
                                
\end{tabular}
\caption{The overall results in single-turn editing.
}
\label{single_turn_main_exp}
\end{table*}

\begin{figure*}[!t]
  \centering
  \includegraphics[width=.95\linewidth]{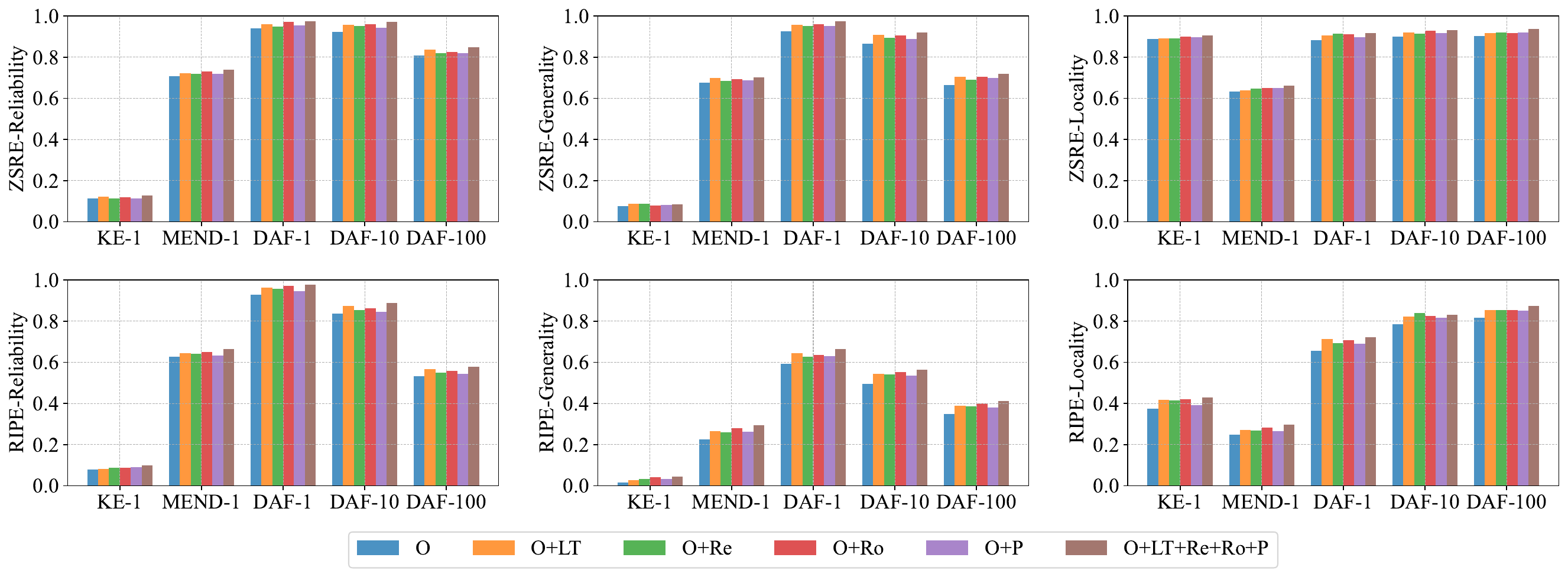} 
  \caption{The influence of different properties in DAFSet for editing results. ``O'', ``LT'', ``Re'', ``P'' and ``Ro'' indicate original, long-tailness, recency, popularity and robustness data, respectively. } 
  \label{training_data_effects}
\end{figure*}

\section{Experiments}
In this section, we extensively evaluate the proposed method and compare it with strong baselines.
Due to space limitation, we describe experimental settings including datasets, baselines and implementation details in Appendix \ref{experiment_settings}.

\subsection{General Results of Sequential Editing}
We first evaluate our DAFNet model over ZSRE \cite{DBLP:conf/conll/LevySCZ17}, CounterFact (CF) \cite{DBLP:conf/nips/MengBAB22} and RIPE datasets \cite{DBLP:journals/corr/abs-2307-12976}, which are editing benchmarks.
We compare the editing performance with 10, 100 and 1000 edits.\footnote{Due to the space limitations, the experimental results of single-turn and 1000 edits in Appendix \ref{editing_data_1k}}

Table \ref{main_exp} shows the overall performance.
We can observe that: (1) Extra-module-based and meta-learning methods show poor results in all metrics. We conjecture that meta-learning methods do not consider the sequential fusion modeling of facts. These methods only focus on  single-turn editing with one fact. KN \cite{DBLP:conf/acl/DaiDHSCW22} achieves editing by amplifying the activation of located neurons, and thus multiple facts can lead to excessive amplification of activation. (2) Locate-then-edit methods in sequential editing scenario show normal performance. The reason may be that multiple back-propagation gradient signals can partially capture the inherent connections between the corresponding sequences. (3) DAFNet achieves significant improvement and the vast majority of results are the best, which proves the effectiveness of our method.

We further evaluate our DAFNet in single-turn editing scenario shown in Table \ref{single_turn_main_exp} with LLAMA2 (7B) backbone.
Our model also achieves the competitive performance over the specifically designed baselines for single-turn model editing.

\subsection{Detailed Analysis}
\noindent\textbf{Influence of Properties in DAFSet.}
We discuss different properties of DAFSet.
We choose typical meta-learning methods such as KE \cite{DBLP:conf/emnlp/CaoAT21} and MEND \cite{DBLP:conf/iclr/MitchellLBFM22} as baselines to compare with DAFNet.
Other baselines do not need extra data to learn random initialized parameters.
We choose the entity-centric ZSRE \cite{DBLP:conf/conll/LevySCZ17} and the relation-centric dataset RIPE \cite{DBLP:journals/corr/abs-2307-12976} as our testing sets.

Based on Figure \ref{training_data_effects}, we can see that (1) For \texttt{Reliability} and \texttt{Generality}, all training data with each property is beneficial for performance improvement on the testing data. 
The KE model shows the performance on individual and combined data is poor. We suggest that the training of KE is based on a fixed batch of editing and modeling on the original model and thus the auxiliary network cannot adapt to new models generated by more than one edit. As the number of edits increases, the performance of these two indicators gradually decreases due to the increase in modeling complexity.
(2) From the \texttt{Locality} results, we can also observe that the model performance can be further improved by adding our data to learn the meta weights. 
The locality results increase as the number of model edits increases.

\noindent\textbf{Influence of Attention Flows.}
In intra- and inter-attention flows, we use the auto-regressive attention mapping to all previous token and fact granularities.
Considering that different window sizes may have different impacts on semantic interaction, we control different window sizes to test the model during editing.
For the inter-attention flow, we need to explore whether the model really pays attention to the previous editing data during the sequence editing process.
Specifically, we evaluate our DAFNet model on ZSRE using LLAMA2.

\begin{figure}[!tb]
  \centering
  \includegraphics[width=8cm,height=6cm]{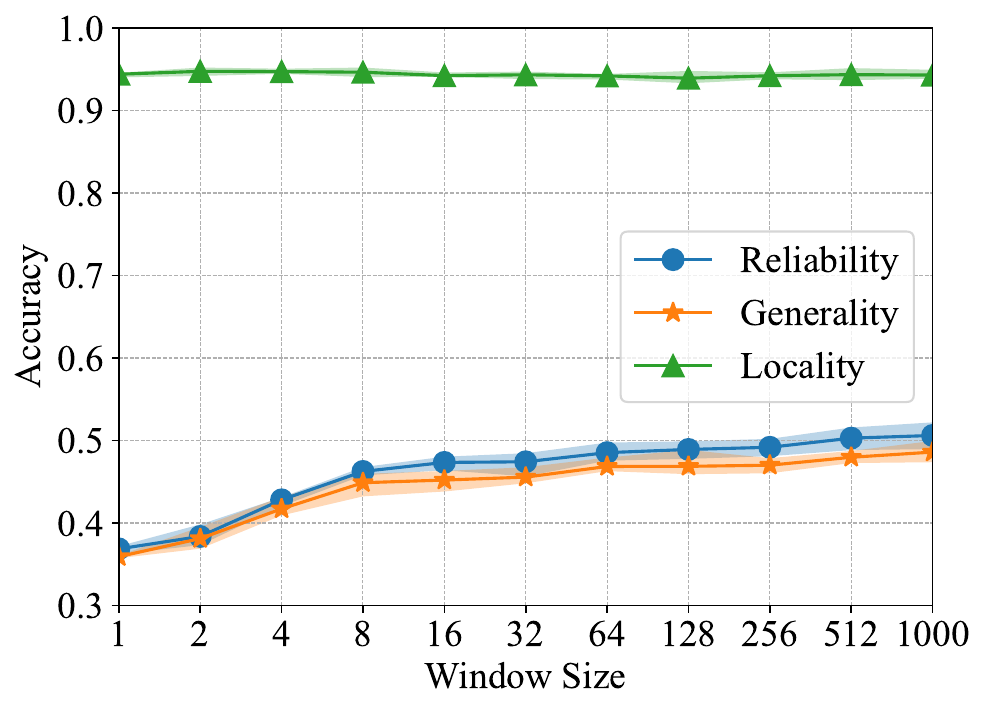} 
  \caption{The  window size influence of intra-editing attention flow.} 
  \label{attention_window_exploration}
\end{figure}

\begin{figure}[!tb]
  \centering
  \includegraphics[width=\linewidth]{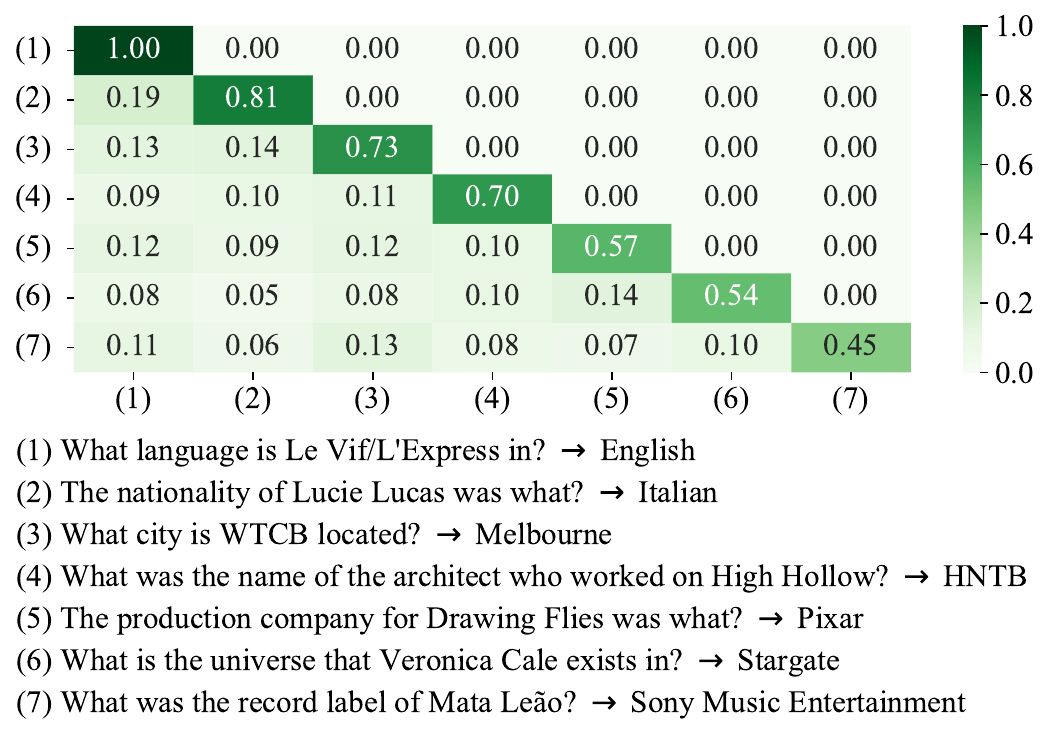} 
  \caption{The inter-editing attention flow scores of 1000 sequential edits.}
  \label{attention_visualization}
\end{figure}

From Figure \ref{attention_window_exploration}, we explore the window size influence of intra-editing attention flow. We can see that as the number of attention modeling window size increases, the results of these two indicators steadily improve.
The reason is that these two editing metrics are highly relevant to the data of the test edited facts and thus the performance can be greatly improved based on the generality of the data.
From Figure \ref{attention_visualization}, the multi-layer inter-editing attention flow shows the self-attention importance with 1,000 edits. We can see that the importance of semantic modeling for each interactive sentence is the greatest, which is consistent with the self-attention mechanism. With the increase of editing interactions, we can see that the data distribution of self-attention becomes more uniform, and the model can better focus on previously edited data.

\begin{table}[!tb]
    \footnotesize
    \setlength{\tabcolsep}{4.1pt}
    \begin{tabular}{ccccccccc}
    \hline
    \multirow{2}{*}{\textbf{Data}} & \multirow{2}{*}{\textbf{intra}} & \multirow{2}{*}{\textbf{Inter}} & \multicolumn{3}{c}{\textbf{10}}      & \multicolumn{3}{c}{\textbf{100}}     \\
                                   &                              &                              & \textbf{R} & \textbf{G} & \textbf{L} & \textbf{R} & \textbf{G} & \textbf{L} \\ \hline
    \multirow{4}{*}{ZSRE}     
    &&&98.5&95.7&93.5&81.7&68.7&89.9\\
    &\checkmark &&99.5&97.2&93.7&88.7&75.1&90.1\\
    &&\checkmark&99.2&96.5&93.8&83.5&71.3&90.1\\
    &\checkmark &\checkmark&\textbf{99.8}&\textbf{98.0}&\textbf{94.3}&\textbf{90.1}&\textbf{76.8}&\textbf{90.3}\\
    \hline
    \multirow{4}{*}{RIPE}   
    &&&80.7&44.6&45.9&54.7&29.1&45.5\\
    &\checkmark &&94.0&56.2&55.8&70.5&37.3&60.5\\
    &&\checkmark&82.6&47.8&48.7&58.1&31.7&52.3\\
    &\checkmark &\checkmark&\textbf{99.2}&\textbf{59.8}&\textbf{58.6}&\textbf{76.3}&\textbf{40.6}&\textbf{64.1}\\
    \hline
    \end{tabular}
    \caption{Ablation study for GPT-J on ZSRE and RIPE. ``R, G, L'' means the three test editing metrics.}
    \label{ablation_study}
\end{table}

\subsection{Ablation Study}
To assess the impact of  intra- and inter-editing attention flows, we conduct an ablation study where we remove each module. This study demonstrates the performance on ZSRE \cite{DBLP:conf/emnlp/YaoWT0LDC023} and RIPE \cite{DBLP:journals/corr/abs-2307-12976} using GPT-J.
From Table \ref{ablation_study}, when the intra-editing attention flow module is removed, the editing performance of LLMs deteriorates rapidly.
The results suffer when the dynamic token modeling process is removed, as it hampers the effective transfer of previously edited semantic memory to the editing of the next token in the facts.
When we remove two main modules simultaneously, our model degenerates to only use the autoregressive attention modeling to perform the input gradient signal. Hence, it is easier to capture the semantic connections between sequentially edited data than other meta learning-based methods.

\section{Conclusion}
In this paper, we propose a dynamic auxiliary fusion model (DAFNet) for sequential editing, including the intra-editing and inter-editing attention flow modules. 
To obtain better meta weights for updating LLMs' original weights in the auxiliary network, we further propose the DAFSet dataset to enhance the editing ability of LLMs.
Experimental results show that our model achieves state-of-the-art results over the strong baselines.

\section*{Limitations}
For DAFSet, the raw data we currently use only contains factual knowledge from the Wiki. Therefore, in the future, we will consider building a more widely distributed training set on other types of raw data.
For DAFNet, due to the need for training, the preparation work before editing is more tedious and time-consuming compared to locate-then-editing based methods. In addition,
due to limitations in machine resources, our model has only been tested at a parameter scale of around 10B. If there are more resources, we can experimentally demonstrate our results on a larger parameter scale.

\section*{Acknowledgements}
We would like to thank anonymous reviewers for their valuable comments.
This work is supported by Alibaba Group through Alibaba Research Intern Program.


\bibliography{custom}

\begin{thebibliography}{36}
\expandafter\ifx\csname natexlab\endcsname\relax\def\natexlab#1{#1}\fi

\bibitem[{Agarap(2018)}]{agarap2018deep}
Abien~Fred Agarap. 2018.
\newblock Deep learning using rectified linear units (relu).
\newblock \emph{arXiv preprint arXiv:1803.08375}.

\bibitem[{An et~al.(2023)An, Ma, Lin, Zheng, Lou, and Chen}]{DBLP:journals/corr/abs-2310-20689}
Shengnan An, Zexiong Ma, Zeqi Lin, Nanning Zheng, Jian{-}Guang Lou, and Weizhu Chen. 2023.
\newblock \href {https://doi.org/10.48550/ARXIV.2310.20689} {Learning from mistakes makes {LLM} better reasoner}.
\newblock \emph{CoRR}, abs/2310.20689.

\bibitem[{Basta et~al.(2021)Basta, Costa{-}juss{\`{a}}, and Casas}]{DBLP:journals/nca/BastaCC21}
Christine Basta, Marta~R. Costa{-}juss{\`{a}}, and Noe Casas. 2021.
\newblock \href {https://doi.org/10.1007/S00521-020-05211-Z} {Extensive study on the underlying gender bias in contextualized word embeddings}.
\newblock \emph{Neural Comput. Appl.}, 33(8):3371--3384.

\bibitem[{Bickel et~al.(2007)Bickel, Br{\"{u}}ckner, and Scheffer}]{DBLP:conf/icml/BickelBS07}
Steffen Bickel, Michael Br{\"{u}}ckner, and Tobias Scheffer. 2007.
\newblock \href {https://doi.org/10.1145/1273496.1273507} {Discriminative learning for differing training and test distributions}.
\newblock In \emph{ICML}, volume 227 of \emph{{ACM} International Conference Proceeding Series}, pages 81--88.

\bibitem[{Blevins et~al.(2023)Blevins, Gonen, and Zettlemoyer}]{DBLP:conf/acl/BlevinsGZ23}
Terra Blevins, Hila Gonen, and Luke Zettlemoyer. 2023.
\newblock \href {https://doi.org/10.18653/V1/2023.ACL-LONG.367} {Prompting language models for linguistic structure}.
\newblock In \emph{ACL}, pages 6649--6663.

\bibitem[{Blinova et~al.(2023)Blinova, Zhou, Jaggi, Eickhoff, and Bahrainian}]{DBLP:conf/acl/BlinovaZJEB23}
Sofia Blinova, Xinyu Zhou, Martin Jaggi, Carsten Eickhoff, and Seyed~Ali Bahrainian. 2023.
\newblock \href {https://doi.org/10.18653/V1/2023.ACL-LONG.552} {{SIMSUM:} document-level text simplification via simultaneous summarization}.
\newblock In \emph{ACL}, pages 9927--9944.

\bibitem[{Brown et~al.(2020)Brown, Mann, Ryder, Subbiah, Kaplan, Dhariwal, Neelakantan, Shyam, Sastry, Askell, Agarwal, Herbert{-}Voss, Krueger, Henighan, Child, Ramesh, Ziegler, Wu, Winter, Hesse, Chen, Sigler, Litwin, Gray, Chess, Clark, Berner, McCandlish, Radford, Sutskever, and Amodei}]{DBLP:conf/nips/BrownMRSKDNSSAA20}
Tom~B. Brown, Benjamin Mann, Nick Ryder, Melanie Subbiah, Jared Kaplan, Prafulla Dhariwal, Arvind Neelakantan, Pranav Shyam, Girish Sastry, Amanda Askell, Sandhini Agarwal, Ariel Herbert{-}Voss, Gretchen Krueger, Tom Henighan, Rewon Child, Aditya Ramesh, Daniel~M. Ziegler, Jeffrey Wu, Clemens Winter, Christopher Hesse, Mark Chen, Eric Sigler, Mateusz Litwin, Scott Gray, Benjamin Chess, Jack Clark, Christopher Berner, Sam McCandlish, Alec Radford, Ilya Sutskever, and Dario Amodei. 2020.
\newblock \href {https://proceedings.neurips.cc/paper/2020/hash/1457c0d6bfcb4967418bfb8ac142f64a-Abstract.html} {Language models are few-shot learners}.
\newblock In \emph{NeurIPS}.

\bibitem[{Cai et~al.(2023)Cai, Namkoong, and Yadlowsky}]{DBLP:journals/corr/abs-2303-02011}
Tiffany~Tianhui Cai, Hongseok Namkoong, and Steve Yadlowsky. 2023.
\newblock \href {https://doi.org/10.48550/ARXIV.2303.02011} {Diagnosing model performance under distribution shift}.
\newblock \emph{CoRR}, abs/2303.02011.

\bibitem[{Cao et~al.(2021)Cao, Aziz, and Titov}]{DBLP:conf/emnlp/CaoAT21}
Nicola~De Cao, Wilker Aziz, and Ivan Titov. 2021.
\newblock \href {https://doi.org/10.18653/V1/2021.EMNLP-MAIN.522} {Editing factual knowledge in language models}.
\newblock In \emph{EMNLP}, pages 6491--6506.

\bibitem[{Cohen et~al.(2023)Cohen, Biran, Yoran, Globerson, and Geva}]{DBLP:journals/corr/abs-2307-12976}
Roi Cohen, Eden Biran, Ori Yoran, Amir Globerson, and Mor Geva. 2023.
\newblock \href {https://doi.org/10.48550/ARXIV.2307.12976} {Evaluating the ripple effects of knowledge editing in language models}.
\newblock \emph{CoRR}, abs/2307.12976.

\bibitem[{Dai et~al.(2022)Dai, Dong, Hao, Sui, Chang, and Wei}]{DBLP:conf/acl/DaiDHSCW22}
Damai Dai, Li~Dong, Yaru Hao, Zhifang Sui, Baobao Chang, and Furu Wei. 2022.
\newblock \href {https://doi.org/10.18653/V1/2022.ACL-LONG.581} {Knowledge neurons in pretrained transformers}.
\newblock In \emph{ACL}, pages 8493--8502.

\bibitem[{Devlin et~al.(2019)Devlin, Chang, Lee, and Toutanova}]{DBLP:conf/naacl/DevlinCLT19}
Jacob Devlin, Ming{-}Wei Chang, Kenton Lee, and Kristina Toutanova. 2019.
\newblock \href {https://doi.org/10.18653/V1/N19-1423} {{BERT:} pre-training of deep bidirectional transformers for language understanding}.
\newblock In \emph{NAACL}, pages 4171--4186.

\bibitem[{Dong et~al.(2022)Dong, Dai, Song, Xu, Sui, and Li}]{DBLP:conf/emnlp/DongDSXSL22}
Qingxiu Dong, Damai Dai, Yifan Song, Jingjing Xu, Zhifang Sui, and Lei Li. 2022.
\newblock \href {https://doi.org/10.18653/V1/2022.FINDINGS-EMNLP.438} {Calibrating factual knowledge in pretrained language models}.
\newblock In \emph{EMNLP}, pages 5937--5947.

\bibitem[{Godbole and Jia(2023)}]{DBLP:conf/eacl/GodboleJ23}
Ameya Godbole and Robin Jia. 2023.
\newblock \href {https://doi.org/10.18653/V1/2023.FINDINGS-EACL.71} {Benchmarking long-tail generalization with likelihood splits}.
\newblock In \emph{EACL}, pages 933--953.

\bibitem[{Hartvigsen et~al.(2022)Hartvigsen, Sankaranarayanan, Palangi, Kim, and Ghassemi}]{DBLP:journals/corr/abs-2211-11031}
Thomas Hartvigsen, Swami Sankaranarayanan, Hamid Palangi, Yoon Kim, and Marzyeh Ghassemi. 2022.
\newblock \href {https://doi.org/10.48550/ARXIV.2211.11031} {Aging with {GRACE:} lifelong model editing with discrete key-value adaptors}.
\newblock \emph{CoRR}, abs/2211.11031.

\bibitem[{Huang et~al.(2023)Huang, Shen, Zhang, Zhou, Rong, and Xiong}]{DBLP:conf/iclr/HuangSZZR023}
Zeyu Huang, Yikang Shen, Xiaofeng Zhang, Jie Zhou, Wenge Rong, and Zhang Xiong. 2023.
\newblock \href {https://openreview.net/pdf?id=4oYUGeGBPm} {Transformer-patcher: One mistake worth one neuron}.
\newblock In \emph{ICLR}.

\bibitem[{Kandpal et~al.(2023)Kandpal, Deng, Roberts, Wallace, and Raffel}]{DBLP:conf/icml/KandpalDRWR23}
Nikhil Kandpal, Haikang Deng, Adam Roberts, Eric Wallace, and Colin Raffel. 2023.
\newblock \href {https://proceedings.mlr.press/v202/kandpal23a.html} {Large language models struggle to learn long-tail knowledge}.
\newblock In \emph{ICML}, volume 202, pages 15696--15707.

\bibitem[{Levy et~al.(2017)Levy, Seo, Choi, and Zettlemoyer}]{DBLP:conf/conll/LevySCZ17}
Omer Levy, Minjoon Seo, Eunsol Choi, and Luke Zettlemoyer. 2017.
\newblock \href {https://doi.org/10.18653/V1/K17-1034} {Zero-shot relation extraction via reading comprehension}.
\newblock In \emph{CoNLL}, pages 333--342.

\bibitem[{Lewis et~al.(2020)Lewis, Liu, Goyal, Ghazvininejad, Mohamed, Levy, Stoyanov, and Zettlemoyer}]{DBLP:conf/acl/LewisLGGMLSZ20}
Mike Lewis, Yinhan Liu, Naman Goyal, Marjan Ghazvininejad, Abdelrahman Mohamed, Omer Levy, Veselin Stoyanov, and Luke Zettlemoyer. 2020.
\newblock \href {https://doi.org/10.18653/V1/2020.ACL-MAIN.703} {{BART:} denoising sequence-to-sequence pre-training for natural language generation, translation, and comprehension}.
\newblock In \emph{ACL}, pages 7871--7880.

\bibitem[{Li et~al.(2022)Li, Gao, Goenka, and Chen}]{DBLP:conf/acl/LiGGC22}
Huihan Li, Tianyu Gao, Manan Goenka, and Danqi Chen. 2022.
\newblock \href {https://doi.org/10.18653/V1/2022.ACL-LONG.555} {Ditch the gold standard: Re-evaluating conversational question answering}.
\newblock In \emph{ACL}, pages 8074--8085.

\bibitem[{Madaan et~al.(2022)Madaan, Tandon, Clark, and Yang}]{DBLP:conf/emnlp/MadaanTCY22}
Aman Madaan, Niket Tandon, Peter Clark, and Yiming Yang. 2022.
\newblock \href {https://doi.org/10.18653/V1/2022.EMNLP-MAIN.183} {Memory-assisted prompt editing to improve {GPT-3} after deployment}.
\newblock In \emph{EMNLP}, pages 2833--2861.

\bibitem[{Meng et~al.(2022)Meng, Bau, Andonian, and Belinkov}]{DBLP:conf/nips/MengBAB22}
Kevin Meng, David Bau, Alex Andonian, and Yonatan Belinkov. 2022.
\newblock \href {http://papers.nips.cc/paper\_files/paper/2022/hash/6f1d43d5a82a37e89b0665b33bf3a182-Abstract-Conference.html} {Locating and editing factual associations in {GPT}}.
\newblock In \emph{NeurIPS}.

\bibitem[{Meng et~al.(2023)Meng, Sharma, Andonian, Belinkov, and Bau}]{DBLP:conf/iclr/MengSABB23}
Kevin Meng, Arnab~Sen Sharma, Alex~J. Andonian, Yonatan Belinkov, and David Bau. 2023.
\newblock \href {https://openreview.net/pdf?id=MkbcAHIYgyS} {Mass-editing memory in a transformer}.
\newblock In \emph{ICLR}.

\bibitem[{Mitchell et~al.(2022{\natexlab{a}})Mitchell, Lin, Bosselut, Finn, and Manning}]{DBLP:conf/iclr/MitchellLBFM22}
Eric Mitchell, Charles Lin, Antoine Bosselut, Chelsea Finn, and Christopher~D. Manning. 2022{\natexlab{a}}.
\newblock \href {https://openreview.net/forum?id=0DcZxeWfOPt} {Fast model editing at scale}.
\newblock In \emph{ICLR}.

\bibitem[{Mitchell et~al.(2022{\natexlab{b}})Mitchell, Lin, Bosselut, Manning, and Finn}]{DBLP:conf/icml/MitchellLBMF22}
Eric Mitchell, Charles Lin, Antoine Bosselut, Christopher~D. Manning, and Chelsea Finn. 2022{\natexlab{b}}.
\newblock \href {https://proceedings.mlr.press/v162/mitchell22a.html} {Memory-based model editing at scale}.
\newblock In \emph{ICML}, pages 15817--15831.

\bibitem[{Omar et~al.(2022)Omar, Choi, Nyang, and Mohaisen}]{DBLP:journals/access/OmarCNM22}
Marwan Omar, Soohyeon Choi, Daehun Nyang, and David Mohaisen. 2022.
\newblock \href {https://doi.org/10.1109/ACCESS.2022.3197769} {Robust natural language processing: Recent advances, challenges, and future directions}.
\newblock \emph{{IEEE} Access}, 10:86038--86056.

\bibitem[{Roumeliotis and Tselikas(2023)}]{DBLP:journals/fi/RoumeliotisT23}
Konstantinos~I. Roumeliotis and Nikolaos~D. Tselikas. 2023.
\newblock Chatgpt and open-ai models: {A} preliminary review.
\newblock \emph{Future Internet}, 15(6):192.

\bibitem[{Shi et~al.(2023)Shi, Chen, Misra, Scales, Dohan, Chi, Sch{\"{a}}rli, and Zhou}]{DBLP:conf/icml/ShiCMSDCSZ23}
Freda Shi, Xinyun Chen, Kanishka Misra, Nathan Scales, David Dohan, Ed~H. Chi, Nathanael Sch{\"{a}}rli, and Denny Zhou. 2023.
\newblock \href {https://proceedings.mlr.press/v202/shi23a.html} {Large language models can be easily distracted by irrelevant context}.
\newblock In \emph{ICML}, pages 31210--31227.

\bibitem[{Tam et~al.(2023)Tam, Mascarenhas, Zhang, Kwan, Bansal, and Raffel}]{DBLP:conf/acl/TamMZKBR23}
Derek Tam, Anisha Mascarenhas, Shiyue Zhang, Sarah Kwan, Mohit Bansal, and Colin Raffel. 2023.
\newblock \href {https://doi.org/10.18653/V1/2023.FINDINGS-ACL.322} {Evaluating the factual consistency of large language models through news summarization}.
\newblock In \emph{ACL}, pages 5220--5255.

\bibitem[{Tan et~al.(2023)Tan, Zhang, and Fu}]{DBLP:journals/corr/abs-2311-04661}
Chenmien Tan, Ge~Zhang, and Jie Fu. 2023.
\newblock \href {https://doi.org/10.48550/ARXIV.2311.04661} {Massive editing for large language models via meta learning}.
\newblock \emph{CoRR}, abs/2311.04661.

\bibitem[{Touvron et~al.(2023)Touvron, Lavril, Izacard, Martinet, Lachaux, Lacroix, Rozi{\`{e}}re, Goyal, Hambro, Azhar, Rodriguez, Joulin, Grave, and Lample}]{DBLP:journals/corr/abs-2302-13971}
Hugo Touvron, Thibaut Lavril, Gautier Izacard, Xavier Martinet, Marie{-}Anne Lachaux, Timoth{\'{e}}e Lacroix, Baptiste Rozi{\`{e}}re, Naman Goyal, Eric Hambro, Faisal Azhar, Aur{\'{e}}lien Rodriguez, Armand Joulin, Edouard Grave, and Guillaume Lample. 2023.
\newblock \href {https://doi.org/10.48550/ARXIV.2302.13971} {Llama: Open and efficient foundation language models}.
\newblock \emph{CoRR}, abs/2302.13971.

\bibitem[{Wang et~al.(2023)Wang, Zhang, Xie, Yao, Tian, Wang, Xi, Cheng, Liu, Zheng, and Chen}]{DBLP:journals/corr/abs-2308-07269}
Peng Wang, Ningyu Zhang, Xin Xie, Yunzhi Yao, Bozhong Tian, Mengru Wang, Zekun Xi, Siyuan Cheng, Kangwei Liu, Guozhou Zheng, and Huajun Chen. 2023.
\newblock \href {https://doi.org/10.48550/ARXIV.2308.07269} {Easyedit: An easy-to-use knowledge editing framework for large language models}.
\newblock \emph{CoRR}, abs/2308.07269.

\bibitem[{Yao et~al.(2023)Yao, Wang, Tian, Cheng, Li, Deng, Chen, and Zhang}]{DBLP:conf/emnlp/YaoWT0LDC023}
Yunzhi Yao, Peng Wang, Bozhong Tian, Siyuan Cheng, Zhoubo Li, Shumin Deng, Huajun Chen, and Ningyu Zhang. 2023.
\newblock \href {https://aclanthology.org/2023.emnlp-main.632} {Editing large language models: Problems, methods, and opportunities}.
\newblock In \emph{EMNLP}, pages 10222--10240.

\bibitem[{Zheng et~al.(2023)Zheng, Li, Dong, Fan, Wu, Xu, and Chang}]{DBLP:journals/corr/abs-2305-12740}
Ce~Zheng, Lei Li, Qingxiu Dong, Yuxuan Fan, Zhiyong Wu, Jingjing Xu, and Baobao Chang. 2023.
\newblock \href {https://doi.org/10.48550/ARXIV.2305.12740} {Can we edit factual knowledge by in-context learning?}
\newblock \emph{CoRR}, abs/2305.12740.

\bibitem[{Zheng et~al.(2022)Zheng, Zhang, Chi, Huang, Tan, Lan, Wei, and Mao}]{DBLP:conf/acl/ZhengZCHTL0M22}
Heqi Zheng, Xiao Zhang, Zewen Chi, Heyan Huang, Yan Tan, Tian Lan, Wei Wei, and Xian{-}Ling Mao. 2022.
\newblock \href {https://doi.org/10.18653/V1/2022.ACL-LONG.288} {Cross-lingual phrase retrieval}.
\newblock In \emph{ACL}, pages 4193--4204.

\bibitem[{Zhong et~al.(2023)Zhong, Wu, Manning, Potts, and Chen}]{DBLP:conf/emnlp/ZhongWMPC23}
Zexuan Zhong, Zhengxuan Wu, Christopher~D. Manning, Christopher Potts, and Danqi Chen. 2023.
\newblock \href {https://aclanthology.org/2023.emnlp-main.971} {Mquake: Assessing knowledge editing in language models via multi-hop questions}.
\newblock In \emph{EMNLP}, pages 15686--15702.

\end{thebibliography}

\appendix

\section{Experimental Settings}
\label{experiment_settings}

\subsection{Training Data}
Following~\citet{DBLP:conf/emnlp/YaoWT0LDC023}, we use the ZSRE training data containing 162555 entries, the CF training data containing 10000 entries, and our proposed enhanced dataset DAFSet to train meta-learning based models, including KE \cite{DBLP:conf/emnlp/CaoAT21}, MEND \cite{DBLP:conf/iclr/MitchellLBFM22}, and our DAFNet model.

\subsection{Evaluation Data} 
\label{data_describe}

\noindent\textbf{ZSRE} \cite{DBLP:conf/conll/LevySCZ17}: It uses BART \cite{DBLP:conf/acl/LewisLGGMLSZ20} to answer questions and manually filtering, where each piece of data contains an editing sample, rephrased counterpart and an irrelevant sample corresponding to the reliability, generality and locality indicators, respectively.
Inspired by \cite{DBLP:conf/emnlp/YaoWT0LDC023}, we divide it into a training set and a testing set with 162555 and 19009 entries.

\noindent\textbf{CF} \cite{DBLP:conf/nips/MengBAB22}: The characteristic is that the facts to be edited are all false facts.
Hence, the probability of the model answering correctly before editing is low, thereby increasing the difficulty of editing evaluation.
Similar to ZSRE, each data contains an editing sample, rephrased data and an irrelevant sample.
Following \cite{DBLP:conf/emnlp/YaoWT0LDC023}, both the training and testing sets contain 10000 entries.

\noindent\textbf{RIPE} \cite{DBLP:journals/corr/abs-2307-12976}: It finely divides the generality and locality into multiple parts. The generality includes logical generalization, combination I, combination II, and subject aliasing \cite{DBLP:journals/corr/abs-2307-12976}. The locality includes forgetfulness and relation specificity. It is also an dataset editing false fact like CF, coupled with its fine-grained evaluation making it a difficult and comprehensive dataset. 
After pre-processing, a total of 4388 entries are collected.
 
\subsection{Baselines}
In this work, besides using fine-tuning as the basic baseline, we mainly compare our DAFNet with three types of editing methods:

\noindent\textbf{Adding Additional Parameters}:
T-Patcher \cite{DBLP:conf/iclr/HuangSZZR023} attaches and trains additional neurons in the FFN of the last layer of the model to be edited.
GRACE \cite{DBLP:journals/corr/abs-2211-11031} proposes a General Retrieval Adapters for Continuous Editing (GRACE), which maintains a dictionary like structure to construct new mappings for potential representations that need to be modified.

\noindent\textbf{Locate-then-Edit}:
(1) KN \cite{DBLP:conf/acl/DaiDHSCW22} uses an integral gradient-based method to locate neurons in FFN, achieving editing by amplifying the activation of the located neurons.
(2) ROME \cite{DBLP:conf/nips/MengBAB22} first uses causal mediation analysis to locate the layer that has the greatest impact on the editing sample. They propose Rank One Model Editing (ROME) to modify the FFN weight of the located layer.
(3) MEMIT \cite{DBLP:conf/iclr/MengSABB23} expands the editing scope to multiple layers based on ROME, which improves editing performance and supports batch editing.

\noindent\textbf{Meta learning-Based}:
(1) KE \cite{DBLP:conf/emnlp/CaoAT21} trains a bidirectional LSTM auxiliary network to predict weight updates of the editing samples.
(2) MEND \cite{DBLP:conf/iclr/MitchellLBFM22} trains an MLP to transform the low-rank decomposition of the gradients of the model to be edited with respect to the editing samples, and updates the model with the transformed gradients to achieve editing.

\begin{algorithm}[!tb]
    \caption{Training of DAFNet}
    \small
    \newcommand{\comm}[1]{\textcolor{gray!50}{\textit{#1}}}
    \begin{algorithmic}[1]
        \STATE \textbf{Input:} Language model to be edited $f$, 
        initialized DAFNet $\mathcal{M}$, 
        training set $\mathcal{D}=\left\{\left(x_e^{(i)}, y_e^{(i)}, 
        \left\{x_{g_j}^{(i)}, y_{g_j}^{(i)}\right\}_{j=1}^{N_g^{(i)}},
        \left\{x_{l_j}^{(i)}\right\}_{j=1}^{N_l^{(i)}}\right)\right\}_{i=1}^N$,
        maximum number of sequential editing modeling $T_{max}$, 
        EMA loss coefficient $\alpha$, EMA loss initial value $L_{ini}$, 
        iteration number to increase sequential modeling number $I_{inc}$, 
        increment scale $\gamma$ for increasing sequential editing modeling number, 
        maximum iteration number  $I_{max}$,
        DAFNet learning rate $\eta$.
        \STATE \textbf{Output:} trained DAFNet $\mathcal{M}$.
        \STATE \comm{\# Current sequential editing modeling number.}
        \STATE $T_{now} = 1$ 
        \STATE \comm{\# Initialize EMA loss and minimum EMA loss.}
        \STATE $L_{min} = L_{ema} = L_{ini}$ 
        \STATE \comm{\# Set the iteration number of minimum EMA loss.}
        \STATE $i_{min} = 1$ 
        \STATE \comm{\# Training iterations for DAFNet $\mathcal{M}$.}
        \FOR{$i \gets 1$ to $I_{max}$}
            \STATE \comm{\# Randomly sample editing number $T$ smaller than current modeling number $T_{now}$.}
            \STATE Sample integer $T$ from $\mathbf{Uniform}(1, T_{now})$
            \STATE $D\leftarrow$ Sample $T$ data from $\mathcal{D}$
            \STATE \comm{\# Collect editing signals for $T$ editing samples.}
            \STATE $S_{edit} = []$  
            \FOR{$(x_e^{(t)}, y_e^{(t)}, \_, \_)$ in $D$}
                \STATE \comm{\# Get editing signal by hook functions.}
                \STATE$u_t, \delta_t = \mathbf{Hook}(f, (x_e^{(t)}, y_e^{(t)}))$
                \STATE $S_{edit}.\mathbf{append}([u_t; \delta_t])$
            \ENDFOR
            \STATE \comm{\# Input editing signals of the $T$ sequential editing modeling samples and obtain corresponding editing weights.}
            \STATE $ [\Delta W_{1},..., \Delta W_{T}], [\Bar{h}_1,...,\Bar{h}_T], \Bar{\beta} = \mathcal{M}\left(S_{edit}, []\right)$ 
            \STATE \comm{\# Compute the fused editing weight of the $T$ editing samples and update the language model $f$.}
            \STATE Compute  $\Delta \tilde{W}_T$ by formula \ref{formula_updated_weights_summation}
            \STATE $f_T = \Gamma(f, \Delta \tilde{W}_T)$
            \STATE \comm{\# Compute loss using data from $D$.}
            \STATE $\mathcal{L}_{total} = \mathcal{L}_{rel}(f_T) + \mathcal{L}_{gen}(f_T) + \mathcal{L}_{loc}(f,f_T)$
            \STATE \comm{\# Update DAFNet $\mathcal{M}$.}
            \STATE $\mathcal{M}\leftarrow\operatorname{Adam}\left(\nabla_{\mathcal{M}} \mathcal{L}_{total}, \eta\right)$
            \STATE \comm{\# Update EMA loss and minimum EMA loss.}
            \STATE $L_{ema} = (1-\alpha) L_{ema} + \alpha \mathcal{L}_{total}$
            \IF{$L_{ema} < L_{min}$}
                \STATE $L_{min} = L_{ema}$
                \STATE $i_{min} = i$ 
            \ENDIF
            \STATE \comm{\# Update current sequential modeling number $T_{now}$, which will increase exponentially until $T_{max}$.}
            \IF{$i-i_{min}>I_{inc}$ and $T_{now}<T_{max}$}
                \STATE $T_{now} = T_{now} + \max(10, \lfloor\gamma T_{now}\rfloor)$
                \STATE $L_{min} = L_{ema} = L_{ini}$
                \STATE $i_{min} = i$
            \ENDIF 
        \ENDFOR
        \RETURN $\mathcal{M}$
    \end{algorithmic}
    \label{alg_dafnet_train}
\end{algorithm}

\begin{algorithm}[!tb]
    \caption{The $t_{th}$ Edit of DAFNet in Sequential Editing Scenario}
    \newcommand{\comm}[1]{\textcolor{gray!50}{\textit{#1}}}
    \small
    \begin{algorithmic}[1]
        \STATE \textbf{Input:} Language model to be edited $f$, 
        trained DAFNet $\mathcal{M}$, 
        the $t_{th}$ edit sample $(x_e^{(t)}, y_e^{(t)})$, 
        the editing weight of previous $t-1$ edits $\Delta \tilde{W}_{t-1}$ 
        (zero when $t=1$),
        the fused fact representations of previous $t-1$ edits 
        $\Bar{H}_{t-1}=[\Bar{h}_1,...,\Bar{h}_{t-1}]$
        (empty list then $t=1$).
        \STATE \textbf{Output:} Edited model $f_t$, the updated 
        editing weights $\Delta \tilde{W}_{t}$, the updated fact representations $\Bar{H}_t$.
        \STATE \comm{\# Get editing signal by hook functions.}
        \STATE$u_t, \delta_t = \mathbf{Hook}(f, (x_e^{(t)}, y_e^{(t)}))$
        \STATE\comm{\# Below input current editing signal $[u_t; \delta_t]$, and past fused fact representations for Intra-editing Attention $\Bar{H}_{t-1}$. 
        Output the editing weight $\Delta W_t$ of the current fact,  
        the fused representations $\Bar{h}_t$ of current fact, and the vector $\Bar{\beta}\in\mathbb{R}^{t}$ described in subsection \ref{subsection_inter_attention_flow}.}
        \STATE$\Delta W_t, \Bar{h}_t, \Bar{\beta} = \mathcal{M}\left([u_t; \delta_t], \Bar{H}_{t-1}\right)$ 
        \STATE \comm{\# Update editing weight.}
        \STATE$\Delta \tilde{W}_t = (1-\Bar{\beta}_{t})\Delta \tilde{W}_{t-1} + \beta_t\Delta W_t$
        \STATE \comm{\# Add updated editing weight to $f$.}
        \STATE $f_t=\operatorname{\Gamma}(f,\Delta \tilde{W}_t)$ 
        \STATE \comm{\# Append fused representation of current fact into the list.}
        \STATE$\Bar{H}_t = [\Bar{h}_1,...,\Bar{h}_{t-1},\Bar{h}_{t}]$
        \RETURN $f_t$, $\Delta \tilde{W}_{t}$, $\Bar{H}_t$
    \end{algorithmic}
    \label{alg_dafnet_edit_once}
\end{algorithm}

We conduct a comprehensive comparison including the methods with additional or without additional data to train the auxiliary network. Some methods require additional data, while others inherently do not require additional data. Each method can be divided into three categories based on the different editing modes: (1) adding extra parameters modules (2) locate-then-editing and (3) meta-learning based approach. Both meta-learning based methods and locate-then-editing based methods require additional data at different stages such as ``ROME, MEMIT, KE, MEND''.
Our DAFSet dataset aims to enhance meta-learning based editing methods. The meta-learning methods in our main experiment of Table \ref{main_exp} are all trained on data enhanced with DAFSet such as ``KE'' and ``MEND''. From Figure \ref{training_data_effects}, we can observe that our DAFNet model also achieves SOTA competitive performance without using additional DAFSet data. If DAFSet data is used, our modeling performance for SME can be further improved.

\subsection{Model Settings and Training Details}

\paragraph{DAFNet}
(1) Hyperparameter Settings: We use GPT-J \footnote{\url{https://huggingface.co/docs/transformers/model_doc/gptj}} and LLAMA2 \footnote{\url{https://huggingface.co/docs/transformers/model_doc/llama2}} as our backbone models to edit and the same hyperparameter settings for the DAFNet auxiliary network. The basic module (including intra-editing attention flow and inter-editing attention flow) of the auxiliary network has 2 layers. 
We set $d_{down}=1024$. All self-attention's head numbers and middle dimensions (including $K, Q, V, O$) are set as 2 and 1024, respectively. 
Regarding the selection of editing weights, we use settings consistent with MEND and KE: GPT-J and LLAMA2 both use the FFN weights of the last three layers of the model.
Different editing matrices with the same shape and a shared DAFNet. Embedding layers are used to remap the representations of editing matrices of different FFN layers inputting into the same DAFNet.

(2) Training Details: As shown in Algorithm \ref{alg_dafnet_train}, we define the initial sequential editing modeling number $T_{now}=1$. The moving coefficient of exponential moving average (EMA) loss $\alpha=0.01$, and set $I_{inc}=1000$. We set the scale to increase the current sequential editing modeling number $T_{now}$ as 0.25, i.e., $\gamma=0.25$. 
The upper limit for the sequential editing modeling number is 1000, i.e.,  $T_{max}=1000$. 
When the sequential editing modeling number reaches the maximum value, we perform an additional 20000 iterations before stopping.
We store checkpoints every 1000 iterations and the checkpoint with the lowest loss would be selected for evaluation. 
The learning rate $\eta$ is set as 1e-6.
The training process takes 7 days on 8 NVIDIA A800 GPUs.
These experiments are presented on average with 5 random runs with different random seeds and the same hyper-parameters.

\paragraph{Baseline Models}
For the baselines, we use the same settings in EasyEdit \cite{DBLP:journals/corr/abs-2308-07269} to train and evaluate other editing methods.

\section{The Editing Algorithm}
In order to facilitate readers to better understand the model training and editing process, we have presented the algorithm pipeline of the training and editing in Algorithm \ref{alg_dafnet_train} and Algorithm \ref{alg_dafnet_edit_once}.

\begin{table*}[tb]
\centering
\small
\includegraphics[width=1\textwidth]{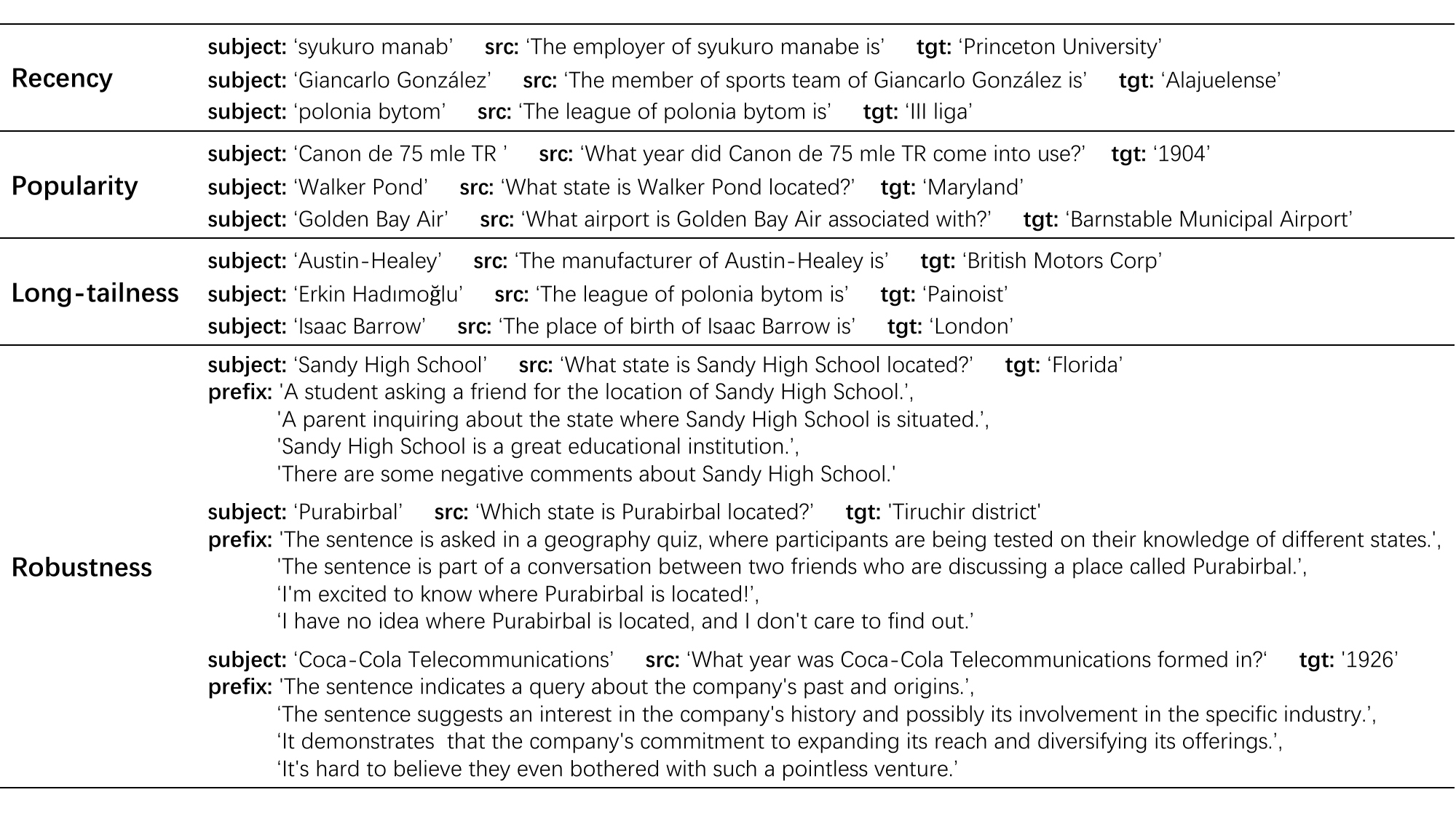} 
\caption{Samples of the DAFSet dataset.}
\label{manual_dataset}
\end{table*}

\begin{table*}[!tb]
\footnotesize
\centering
\begin{tabular}{cccccc}
\midrule
 \multirow{1}{*}{\textbf{Training Type}} & \multirow{1}{*}{\textbf{Model}} & \textbf{Training Time (Day)} & \textbf{GPU Memory (GB)} & \textbf{Inference Time (s)} & \textbf{Avg.} \\ \midrule
                                
\multirow{6}{*}{No Training Before Editing}&FT& N/A &\textbf{27.80} & 1.73 & 25.09\\
&TP& N/A &32.30&5.56 & 55.63\\
&KN & N/A & 31.50 &20.41 &9.08\\
& ROME & N/A & 36.80 & 16.10 & 56.85 \\
 & MEMIT & N/A & 42.90 & 31.65 & 50.40 \\
& GRACE & N/A & 35.20 & 0.16 & 54.12 \\ \midrule
\multirow{4}{*}{ Training Before Editing}&KE & 3 & 41.10 & 0.26  & 10.53\\
& MEND & 1 & 59.40 & \textbf{0.08} & 19.79 \\
& MALMEN & 2 & 56.20 & 2.18 &52.40 \\
 & DAFNet & 7 & 37.20 & 0.29 & \textbf{80.42} \\
\midrule
                                
\end{tabular}
\caption{The overall comparison of computation efficiency.}
\label{time_memory_trade_off}
\end{table*}


\begin{table*}[!tb]
\scriptsize
\centering
\setlength{\tabcolsep}{4.3pt}
\begin{tabular}{ccccccccccccccc}
\midrule
\multirow{2}{*}{\textbf{Backbone}} & \multirow{2}{*}{\textbf{\# Editing}} & \multirow{2}{*}{\textbf{Editor}} & \multicolumn{4}{c}{\textbf{ZSRE}}                 & \multicolumn{4}{c}{\textbf{CounterFact}}                   & \multicolumn{4}{c}{\textbf{RIPE}}                 \\
                                &                               &                                  & \textbf{Rel.} & \textbf{Gen.} & \textbf{Loc.} & \textbf{Avg.} & \textbf{Rel.} & \textbf{Gen.} & \textbf{Loc.} & \textbf{Avg.} & \textbf{Rel.} & \textbf{Gen.} & \textbf{Loc.} & \textbf{Avg.} \\ \midrule

\multirow{10}{*}{\makecell[c]{GPT-J \\ (6B)} }
&\multirow{10}{*}{1000}&FT&4.3&3.0&0.1&2.5$_{(\pm0.1)}$&12.9&5.1&1.1&6.4$_{(\pm0.1)}$&3.1&0.9&0.8&1.6$_{(\pm0.0)}$\\
&&TP&45.7&40.4&10.5&32.2$_{(\pm0.8)}$&47.3&17.0&1.4&21.9$_{(\pm0.7)}$&48.1&29.1&15.2&30.8$_{(\pm0.6)}$\\
&&KN&0.8&0.0&2.2&1.0$_{(\pm0.0)}$&0.1&0.4&1.0&0.5$_{(\pm0.0)}$&0.0&0.0&0.0&0.0$_{(\pm0.0)}$\\
&&ROME&57.2&53.9&29.9&47.0$_{(\pm1.1)}$&0.2&0.2&0.0&0.1$_{(\pm0.0)}$&47.5&16.9&13.4&26.0$_{(\pm0.5)}$\\
&&MEMIT&56.8&54.6&54.9&55.4$_{(\pm1.3)}$&\textbf{82.3}&36.4&30.7&49.8$_{(\pm1.3)}$&0.0&0.0&0.0&0.0$_{(\pm0.0)}$\\
&&GRACE&56.2&51.3&28.4&45.3$_{(\pm1.2)}$&0.3&0.4&0.1&0.3$_{(\pm0.1)}$&46.7&16.3&13.8&25.6$_{(\pm0.7)}$\\
&&$\text{KE}^\spadesuit$&0.0&0.0&1.1&0.4$_{(\pm0.0)}$&0.0&0.0&0.1&0.0$_{(\pm0.0)}$&0.0&0.0&0.2&0.1$_{(\pm0.0)}$\\
&&$\text{MEND}^\spadesuit$&0.0&0.0&0.0&0.0$_{(\pm0.0)}$&0.0&0.0&0.0&0.0$_{(\pm0.0)}$&0.2&0.1&0.1&0.1$_{(\pm0.0)}$\\
&&$\text{MALMEN}^\spadesuit$&43.0&35.1&39.3&39.1$_{(\pm0.4)}$&15.0&12.4&25.1&17.5$_{(\pm0.4)}$&31.1&19.1&35.3&28.5$_{(\pm0.6)}$\\
&&$\text{DAFNet}^\spadesuit$&\textbf{60.0}&\textbf{57.6}&\textbf{88.0}&\textbf{68.5}$_{(\pm1.9)}$&53.1&\textbf{38.1}&\textbf{82.3}&\textbf{57.8}$_{(\pm1.2)}$&\textbf{48.3}&\textbf{31.3}&\textbf{57.3}&\textbf{45.6}$_{(\pm1.2)}$\\
\midrule

\multirow{10}{*}{\makecell[c]{LLAMA2 \\ (7B)}}
&\multirow{10}{*}{1000}&FT&7.9&6.7&4.6&6.4$_{(\pm0.1)}$&1.5&0.1&1.7&1.1$_{(\pm0.0)}$&2.7&1.0&2.2&2.0$_{(\pm0.0)}$\\
&&TP&47.7&44.1&4.4&32.0$_{(\pm0.6)}$&\textbf{64.7}&32.5&11.6&36.3$_{(\pm0.9)}$&42.3&26.8&9.9&26.3$_{(\pm0.6)}$\\
&&KN&0.0&0.0&0.0&0.0$_{(\pm0.0)}$&0.0&0.0&0.0&0.0$_{(\pm0.0)}$&0.0&0.0&0.0&0.0$_{(\pm0.0)}$\\
&&ROME&1.6&1.5&0.6&1.2$_{(\pm0.0)}$&0.2&0.1&0.1&0.1$_{(\pm0.0)}$&0.0&0.0&0.0&0.0$_{(\pm0.0)}$\\
&&MEMIT&0.2&0.2&0.1&0.2$_{(\pm0.0)}$&0.1&0.1&1.0&0.4$_{(\pm0.0)}$&0.0&0.0&0.0&0.0$_{(\pm0.0)}$\\
&&GRACE&1.5&1.6&0.8&1.3$_{(\pm0.1)}$&0.1&0.2&0.1&0.1$_{(\pm0.0)}$&0.0&0.0&0.0&0.0$_{(\pm0.0)}$\\
&&$\text{KE}^\spadesuit$&0.0&0.0&0.0&0.0$_{(\pm0.0)}$&0.0&0.0&0.3&0.1$_{(\pm0.0)}$&0.0&0.0&0.0&0.0$_{(\pm0.0)}$\\
&&$\text{MEND}^\spadesuit$&0.0&0.0&0.0&0.0$_{(\pm0.0)}$&0.0&0.0&0.0&0.0$_{(\pm0.0)}$&0.0&0.0&0.0&0.0$_{(\pm0.0)}$\\
&&$\text{MALMEN}^\spadesuit$&32.0&28.5&28.1&29.6$_{(\pm0.6)}$&15.8&16.4&22.5&18.3$_{(\pm0.3)}$&42.3&38.4&38.5&39.8$_{(\pm0.9)}$\\
&&$\text{DAFNet}^\spadesuit$&\textbf{50.5}&\textbf{48.6}&\textbf{93.6}&\textbf{64.2}$_{(\pm1.3)}$&50.4&\textbf{35.8}&\textbf{76.9}&\textbf{54.4}$_{(\pm1.6)}$&\textbf{44.1}&\textbf{34.3}&\textbf{85.8}&\textbf{54.7}$_{(\pm0.9)}$\\
\midrule
                                
\end{tabular}
\caption{Results with 1000 edits of DAFNet and baselines.}
\label{edits_res_1k}
\end{table*}

\section{Additional Experimental Results}
\subsection{Dataset Construction}
\label{appendix_dafset}
The collected dataset samples are shown in Table \ref{manual_dataset}, including the manual templates used to prompt the LLMs.
The detailed long-tail dataset construction process is shown as follows:
\begin{itemize}
    \item We use the LLMs to perform language modeling on each data in the dataset to obtain the log likelihood probability for each token position in the sentence.
    \item The sentence semantic representation is obtained by multiplying the log likelihood probability of all tokens.
    \item We select the subject in the sentence that is lower than the threshold as the long tail sentence.
\end{itemize}
Since the editing samples are all obtained through a single triple transformation, each editing sample only contains one entity and relation such as above samples. Therefore, the semantic of this sentence is usually dominated by entity and corresponding relation. If the log likelihood probability of the sentence is relatively low, it indicates that the semantic of the entity triple is not well memorized in the LLMs and the entity triple is low-frequency sparse knowledge \cite{DBLP:conf/eacl/GodboleJ23,DBLP:conf/icml/KandpalDRWR23}. Hence, it can be used for capturing the long-tailnesses of those subject entities.
Note that the construction process of long-tail data involves concatenating knowledge triples with conjunctions or articles to form a natural language. Their training samples lengths are basically the same.
Meanwhile, we multiply by the corresponding sample length to maintain the fairness of the prediction probability as much as possible.
Most samples have lengths between 6 and 8, and thus there is no unfairness in comparison after multiplication.

\subsection{Computational Resource Analysis}
In order to evaluate the computational cost of our model and baselines, we compare the scores of different editors including ``No training'' and ``Training'' before editing on all datasets. The average score is performed on the editing numbers of 1, 10 and 100.
Specifically, we evaluate the model's overhead on machine resources by comparing Training time, GPU memory and Inference time. 

From the Table \ref{time_memory_trade_off}, we can conclude that although our DAFNet model is not the optimal design in terms of  machine resources overhead, our model results have achieved significant improvement under the GPU memory. Our inference time also has strong competitiveness. Since the LLMs are usually trained once and can be reused, our training time is also acceptable.
The high memory overhead is mainly due to the need for two different auxiliary networks to model the semantic interaction within and between facts.
We can unify the modeling of the fusion process in two scenarios to save memory costs in the future.

\subsection{Results of 1000 Sequential Edits}
\label{editing_data_1k}
The results of 1000 sequential edits are presented in Table \ref{edits_res_1k}. They also show the similar conclusion with general results, which prove the effectiveness of our approach.

\end{document}